%% file: corl_2026_template_submission/main.tex
\documentclass{article}

\usepackage[final]{corl_2026} 
\usepackage[dvipsnames,table]{xcolor}     
\usepackage{colortbl}
\usepackage{siunitx}
\usepackage{amsmath, amssymb, bm}   
\usepackage{graphicx, subcaption}   
\usepackage{booktabs, multirow}     
\usepackage{algorithm, algpseudocode} 
\usepackage{hyperref}     
\hypersetup{
    colorlinks=true,       
    urlcolor=MidnightBlue, 
    linkcolor=BrickRed,    
    citecolor=ForestGreen,  
    pdftitle={Contact-UAN: Unsupervised Actuator Networks as a Residual World Model for Humanoid Control},
    pdfauthor={Nolan Fey, Khai Nguyen, Yash Prabhu, Gabriel B. Margolis, Pulkit Agrawal},
    pdfkeywords={Sim-to-Real Transfer, Humanoids, Residual World Models}
}
\usepackage[capitalize]{cleveref} 
\usepackage{tikz, pgfplots, pgfplotstable} 
\pgfplotsset{compat=1.18}
\definecolor{BaselineGray}{HTML}{6F6F6F}
\definecolor{UANTeal}{HTML}{0072B2}
\newsavebox{\openlooprolloutsbox}

\title{Contact-UAN: Unsupervised Actuator Networks as a \\ Residual World Model for Humanoid Control}

\author{
Nolan Fey$^{*}$, 
Khai Nguyen$^{*}$,
Yash Prabhu,
Gabriel B. Margolis,
Pulkit Agrawal
\\
Improbable AI Lab, MIT, $^*$Co-first authors.
\\
\texttt{\{nolanfey,khain,yprabhu,gmargo,pulkitag\}@mit.edu}
\\[0.5em]
\textbf{\url{http://contact-uan.csail.mit.edu/}}
}

\begin{document}
\maketitle

\begin{minipage}{\linewidth}
    \centering
    \includegraphics[width=\linewidth, trim=0cm 1.6cm 0cm 0cm, clip]{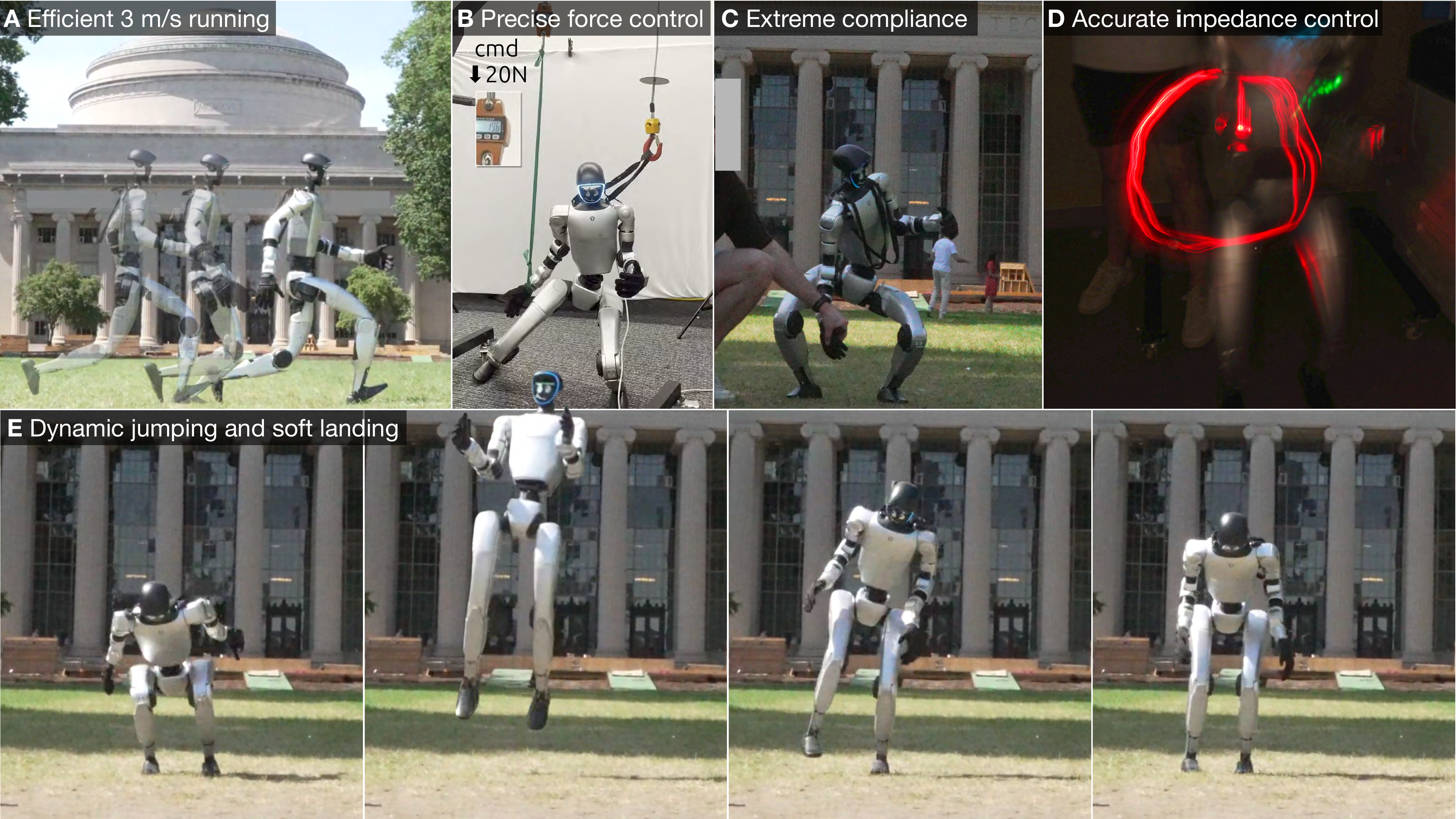}
    \captionof{figure}{\textbf{Sim-to-real transfer of efficient, precise, and dynamic humanoid behaviors.} \emph{Contact-UAN} enables residual world model learning from hardware trajectories with contact by inferring contact-consistent simulator reset states from onboard proprioception. The resulting residual world model constrains policy learning to actuator-feasible behaviors.
    }
    \label{fig:main}
\end{minipage}

\begin{abstract} 
Humanoid policies trained in simulation can transfer qualitatively to hardware while failing to transfer the force interactions learned during training, compromising torque efficiency, load regulation, and robustness to impacts.
This failure can arise from actuator sim-to-real gaps because forces depend on actuator torques.
We introduce \emph{Contact-UAN}, a residual world model that preserves a simulator's physics priors while correcting actuator dynamics to align simulated rollouts with recorded hardware rollouts. For humanoid motion with contact, this requires contact-consistent replay: if hardware playback in simulation begins with misaligned contact wrenches, the residual model learns to correct reset artifacts rather than actuator discrepancies. Using only onboard proprioception, Contact-UAN infers the \emph{simulator-side stance-foot pose} needed to initialize replay with consistent contact forces.
We show that a UAN trained on 6.5 minutes of real-world walking data generalizes to tasks that benefit from accurate force and impulse regulation and produces policies that transfer seamlessly to the real world, yielding 20\% lower running torque, 43\% lower force-tracking error, extreme compliance, accurate impedance control, and dynamic, robust jumps.

\end{abstract}


\section{Introduction}\label{sec:intro}

Humanoid robots can perform diverse tasks by virtue of their general-purpose morphology, and recent advances in learning-based control within physics simulators have enabled increasingly complex behaviors~\cite{sleiman2026zest,liao2025beyondmimic,wu2026perceptivehumanoidparkourchaining,margolis2025softmimic}. 
For sim-to-real humanoid control, qualitative motion transfer is only a coarse measure of success; the robot must also generate force profiles with the magnitude and timing needed to move efficiently, regulate loads, and absorb impacts.
Policy optimization can exploit imperfect actuator dynamics, producing policies that may transfer qualitative motions but not force interactions, leading to excessive torques, poor commanded-force tracking, or failures during high-impulse phases (\cref{sec:sim2real_transfer}).
Because these forces are mediated by joint torques, we study how improving the actuator model used during policy training reduces force-level sim-to-real gaps.

The actuator model fidelity required for sim-to-real transfer is task-dependent: simple models may suffice for quasi-static behaviors, while behaviors such as efficient running, force tracking, jumping, and landing depend on accurate actuator response, force generation, and impact absorption. Analytic simulators~\cite{NVIDIA_Isaac_Sim,mujoco} provide useful physical structure, yet they often approximate actuator effects such as torque saturation and load-dependent nonlinear friction, precisely the effects that shape ground-force generation, external-force regulation, and impact absorption~\cite{Hwangbo_2019,fey2025bridging}. 
Domain randomization improves robustness by training over parameter ranges, but it can produce conservative policies that under-utilize hardware~\cite{tobin2017domain,muratore2022robot,xie2021dynamics,campanaro2024learning}. Online adaptation and SysID can recover performance by estimating dynamics parameters, but require choosing which simulator parameters to randomize and identify~\cite{kumar2021rmarapidmotoradaptation,liu2025locoformer,bjelonic2025towards,liao2025beyondmimic,sleiman2026zest,margolis2022rapidlocomotionreinforcementlearning,margolis2023learningphysicalpropertiesactive}. 
Fully learned world models (WMs) take another approach: fit dynamics directly from data and can, in principle, capture effects omitted by analytical simulators~\cite{hou2026world,li2025robotic,hansen2025hierarchical,wu2023daydreamer}, but must model the full robot-environment dynamics needed for policy optimization. We therefore follow prior work in residual modeling and actuator learning~\cite{gruenstein2021residual,zeng2020tossingbot,golemo2018sim,gao2024sim,jiang2021simgan,yu2025neural,Hwangbo_2019}: retain the physical priors of an existing simulator and learn only the sim-to-real gap in the actuator command channel.

Learning such an actuator-channel residual to improve downstream transfer requires a training signal. Supervised actuator networks use measured motor torques as this signal~\cite{Hwangbo_2019}, but reliable output torque sensing is unavailable on many humanoid platforms.
Without such labels, residual methods can use trajectory replay-based training instead: reset the simulator to a recorded hardware state, apply the recorded commands, and train residual commands so the simulated rollout matches the recorded hardware trajectory~\cite{fey2025bridging,he2025asap}. Critically, this objective isolates modeling error only if replay begins from a hardware-equivalent simulator state. Otherwise, the residual model may learn to correct simulation reset artifacts rather than underlying actuator discrepancies.

\begin{figure}[t]
    \centering
    \begin{minipage}[t]{0.66\linewidth}
        \centering
        \includegraphics[width=\linewidth]{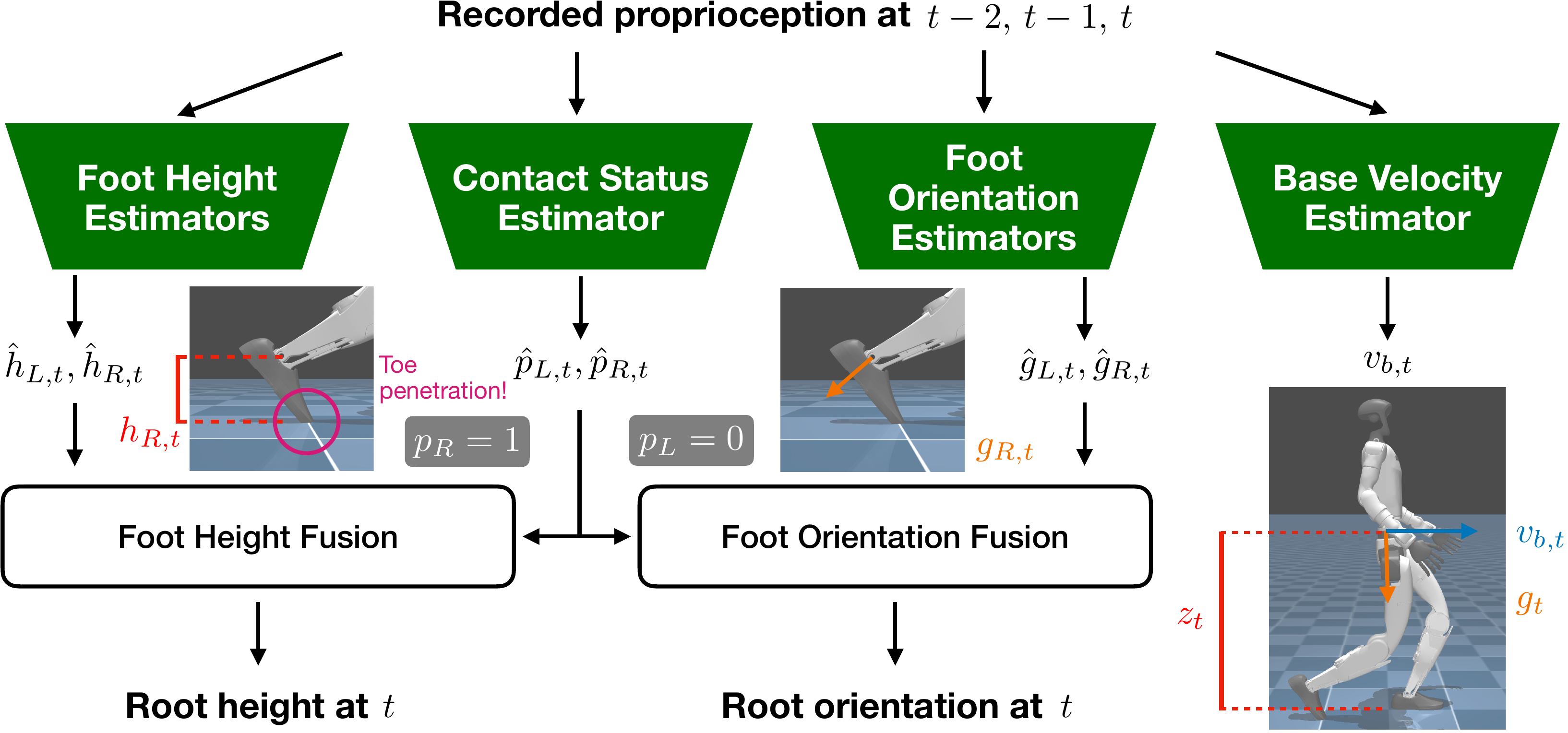}
        \captionof{figure}{\textbf{Our proprioceptive state estimation framework.} Simulation supervision maps load-dependent proprioceptive cues to the simulator-side stance-foot pose consistent with them under the soft-contact model. Equations ~\eqref{eq:contact_weights}--\eqref{eq:contact_weighted_root_orientation} fuse confident contacts into the root pose used for hardware replay in simulation. 
        }
        \label{fig:state_estimation_block_diagram}
    \end{minipage}
    \hfill
    \begin{minipage}[t]{0.32\linewidth}
        \centering
        \includegraphics[width=\linewidth]{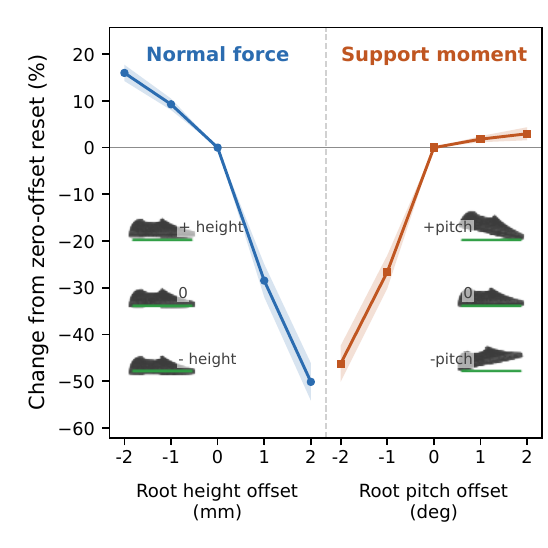}
        \captionof{figure}{\textbf{Contact-wrench sensitivity.} A \SI{}{\milli\meter}-scale height error substantially changes force, and a degree-scale pitch error substantially changes moment.}
        \label{fig:contact_pose_sensitivity}
    \end{minipage}
    \vspace{-1em}
\end{figure}

Matching a given hardware state in simulation is a kinematics problem for suspended, free-space motions~\cite{fey2025bridging,bjelonic2025towards}. However, such motion lacks the contact information needed to capture load-dependent actuator behavior during stance, force generation, impact, and landing.
\citet{he2025asap} collect data with contacts and use a motion capture system to provide ground-truth body poses. However, consider a humanoid robot standing still: on hardware, ground-reaction forces balance its weight, but initializing a \emph{soft-contact} simulator with the feet exactly at the ground surface may not produce equivalent forces. Instead, matching the hardware force interaction requires introducing slight ground penetration via adjustments to the simulated foot height and orientation, not a precise height measurement. 
This is necessary because millimeter-scale foot-height and degree-scale orientation errors can substantially change simulated contact wrenches (\cref{fig:contact_pose_sensitivity}).
These observations point to a critical need to address contact-consistent state initialization for replay-based residual learning.

We introduce \emph{Contact-UAN}, a residual world-modeling approach to training humanoid policies for force- and impulse-sensitive behaviors. The core contribution is contact-consistent state initialization from proprioception, which enables replay-based residual learning from hardware trajectories with contacts. From proprioception alone, Contact-UAN infers the simulator-side stance-foot pose, including the small ground penetration and tilt needed to initialize replay with contact wrenches consistent with the hardware trajectory, together with the base velocity needed to complete the reset state. With replay initialized in this contact-consistent state, the UAN residual can learn to correct true actuator modeling errors in simulation.

We evaluate Contact-UAN on a Unitree G1 using the state-of-the-art BeyondMimic G1 parameters as the nominal simulator~\cite{zakka2026mjlab,liao2025beyondmimic}, and compare policies trained with and without UAN (\cref{sec:result}). We choose downstream tasks that expose distinct actuator-sensitive force and impulse regimes: running requires torque-efficient ground-force generation, force and impedance control require accurate external-force regulation, and jumping requires takeoff impulse and landing absorption. A single UAN trained from 6.5 minutes of real-world walking data improves transfer across these behaviors. 

Training in this corrected simulator yields a running policy that preserves velocity tracking while reducing hardware torque usage by 20\% relative to the no-UAN baseline, showing that the UAN gives policy learning a more accurate torque signal. The same residual reduces force-tracking error by 43\% and improves jumping: both reference-tracking policies transfer, while task-reward fine-tuning for higher jumps transfers with UAN but fails without it (\cref{sec:sim2real_transfer}). Sim-to-sim experiments reveal the importance of contact-consistent initialization and show that Contact-UAN remains effective under larger actuator gaps (\cref{sec:residual_accuracy,sec:sim2sim_transfer}).

\section{Method}\label{sec:method}
\begin{figure}[tbp]
    \centering
    \includegraphics[width=0.9\linewidth]{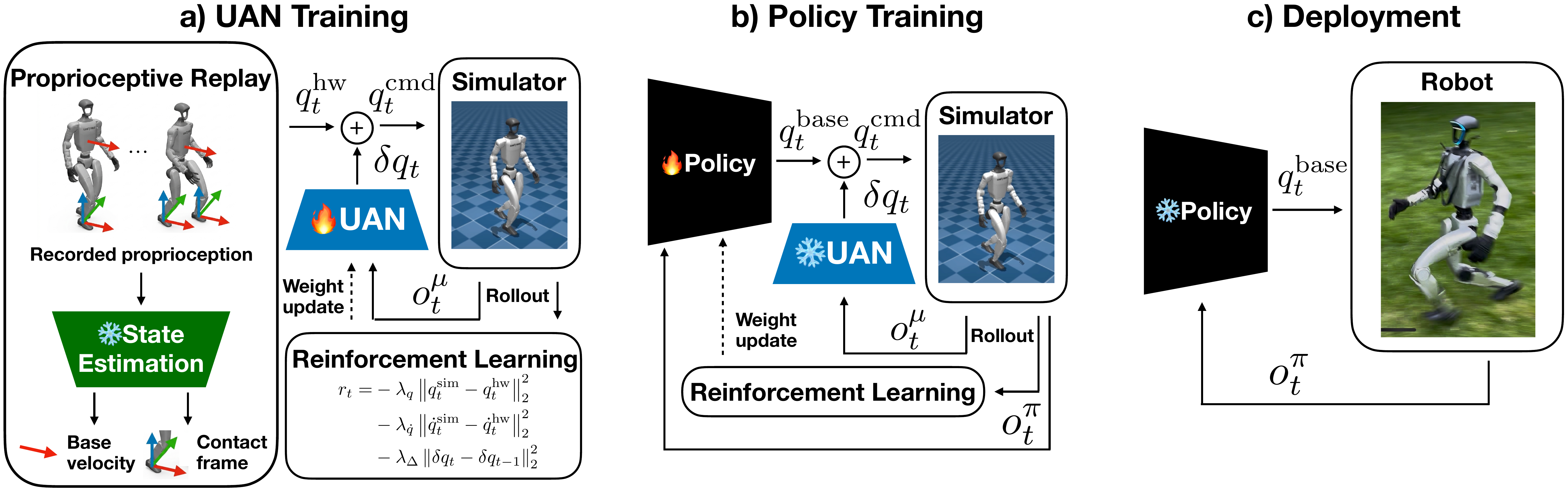}
    \caption{\textbf{Contact-UAN.} a) Train a UAN to add joint-position target corrections to reduce simulation transition error relative to recorded hardware data, with base velocities and contact poses labeled via proprioceptive state estimators. b) Train a task policy in the UAN-augmented simulator with no robot-specific domain randomization. c) Deploy only the learned policy on hardware.}
    \label{fig:block_diagram}
    \vspace{-1em}
\end{figure}

\emph{Contact-UAN} has three stages (\cref{fig:block_diagram}): relabel hardware
data for contact-consistent replay initialization (\cref{ssec:state_estimation}); train
the UAN residual (\cref{sec:uan_training}); and train downstream policies in the augmented
simulator (\cref{sec:downstream_training}). Only the task policy runs on hardware.

\subsection{Contact-Consistent Replay Initialization}\label{ssec:state_estimation}

Accurate replay requires each window to begin with simulated contacts that reproduce
hardware contact wrenches. Onboard sensing provides neither the contact wrench nor absolute base height, and reconstructing global pose is inadequate because soft-contact wrenches depend on simulator-specific, millimeter-scale foot penetration and tilt (\cref{fig:contact_pose_sensitivity}). We therefore infer the \emph{simulator-side} contact-consistent reset state from proprioception.

Contact affects joint motion and IMU response, making contact wrenches inferable from a short proprioceptive history~\cite{margolis2025softmimic,portela2024learning}. The state estimators are trained in simulation and assume flat terrain, enabling use of a root-$xy$ and -yaw invariant frame. They predict contact probabilities, stance-foot pose, and base linear velocity from a $K$-step history of IMU measurements, joint states, and previous joint targets. 

Although proprioception could be used to estimate contact force directly, a force estimate alone does not specify the simulator state used at reset. We therefore regress simulator-side stance-foot pose, learning in one step the map from load-dependent proprioception to the penetrating height and orientation that reproduce the contact wrenches. Initialization from this pose avoids separately inverting the simulator's contact model. It is critical that absolute base height is omitted from the observations: otherwise, the estimator can minimize its simulation loss by copying height rather than using load-dependent cues. This shortcut need not transfer because hardware height may differ from the simulator height that reproduces the same contact force.

We first estimate contact status and train the pose regressors only on stance samples. Swing poses largely map to zero contact wrench, so their variation dominates the objective without informing the contact model. In stance, however, small pose differences strongly affect the wrench: in our MuJoCo model (Table~\ref{tab:mujoco_contact_params})~\cite{mujoco}, a $+2$ mm root-height offset reduces normal force by roughly $50\%$, while a $-2^\circ$ pitch offset changes the support moment by roughly $45\%$
(\cref{fig:contact_pose_sensitivity}). Stance conditioning therefore focuses the regression on the contact-sensitive variation needed at reset
(\cref{fig:state_estimation,fig:open_loop_rollout}).

Each confident stance-foot estimate yields a candidate root height and
orientation through forward kinematics. We fuse candidates using
contact-probability weights and discard windows with no confident contact;
full equations are in Appendix~\ref{app:state_estimation}. Because the
estimators are trained in simulation, their key transfer assumption is
that load-dependent proprioceptive cues remain informative across domains.
\Cref{fig:state_estimation}A probes this assumption in controlled sim-to-sim:
the stance estimator, trained in the base simulator, achieves low contact-height
error in the \texttt{hard} target simulator despite actuator parameter perturbations unseen during training.
\Cref{fig:state_estimation}B and Appendix Figure~\ref{fig:knee_position_replay} provide complementary end-to-end evidence
through improved hardware replay. The estimators are used only for offline
relabeling and are absent from downstream policy training and deployment.
Together with estimated base velocity, the fused root pose initializes
contact-consistent UAN replay, enabling the UAN to learn to correct actuator discrepancies rather than simulator reset artifacts.

\subsection{UAN as a Residual World Model}\label{sec:uan_training}
Let $s_t$ denote the robot state and let $a_t$ denote the action. The real robot transition is $f(s_t,a_t)$, while the baseline simulator transition is $f_{\mathrm{sim}}(s_t,a_t)$. We learn a residual model $\mu$ that corrects the simulator input rather than the simulator state:
\begin{equation}
    \min_{\mu}\;
    \mathbb{E}_{(s_t,a_t)}
    \left[
    \left\|
    f(s_t,a_t) -
    f_{\mathrm{sim}}\!\left(s_t,a_t + \mu(o^\mu_t)\right)
    \right\|_2^2
    \right].
    \label{eq:uan_residual_objective}
\end{equation}
In implementation, $\mu$ outputs a residual correction $\delta q_t$ that is added to a base target $q^{\mathrm{base}}_t$:
\begin{equation}
    q^{\mathrm{cmd}}_t = q^{\mathrm{base}}_t + \delta q_t,
    \qquad
    \delta q_t = \mu(o^\mu_t).
    \label{eq:uan_cmd}
\end{equation}
During UAN training, $q^{\mathrm{base}}_t \equiv q^{\mathrm{hw}}_t$ is the recorded replay target, and during downstream policy training, $q^{\mathrm{base}}_t$ is the policy output. The UAN therefore preserves the simulator's rigid-body dynamics, actuator implementation, contact solver, and numerical integration; it only modulates the command that those existing simulation components integrate forward.

\textbf{UAN observations and architecture.}
The UAN observes joint-local tracking signals over a short history. For each joint, the observation history contains the target-position error and joint velocity,
\begin{equation}
    o^\mu_t =
    \left[
    (q^{\mathrm{base}} - q)_{t-H_\mu+1:t},
    \dot{q}_{t-H_\mu+1:t}
    \right].
    \label{eq:uan_obs}
\end{equation}
This compact observation limits the residual model to local actuator-tracking information to avoid overfitting~\cite{Hwangbo_2019,fey2025bridging}. In our experiments we set $H_\mu=3$ (\SI{60}{\milli\second}).

To improve data efficiency, we share network parameters across joints with identical actuator hardware \cite{Hwangbo_2019,fey2025bridging}. Each actuator type has its own two-layer MLP with hidden dimensions $(128,128)$ and ELU activations. This enforces a hardware-structured residual model while having enough capacity to model actuator-specific nonlinearities.

\textbf{Data collection.}
We first train a Walk-These-Ways (WTW)-style locomotion policy~\cite{margolis2023walk}
in the nominal simulator without a UAN, using domain randomization to
support initial hardware transfer. We then deploy this policy to collect
the hardware trajectories used for UAN training, varying gait frequency
and swing height to diversify stance durations, swing speeds, and impacts.
We record onboard proprioception: base angular velocity, projected gravity, joint positions and velocities, and commanded joint targets; then relabel each hardware window with base velocity and pose following Section~\ref{ssec:state_estimation}.
Appendix~\ref{app:sec:wtw} gives WTW policy-training details.

\textbf{UAN task design.}
The UAN is trained in a replay environment built from labeled hardware windows via PPO~\cite{schulman2017proximal}. At reset, each simulated robot is initialized from a randomly sampled replay window. We use a plane terrain, 4096 parallel environments, a 24-step (\SI{480}{\milli\second}) replay horizon, and PPO rollouts of the same length. 
The reward penalizes joint-position MSE, joint-velocity MSE, and residual-rate magnitude. 
This replay task encourages the UAN to learn corrections that make simulated joint transitions match hardware while discouraging unnecessarily high-frequency residual commands to aid with exploration.

\input{figures/camera_ready/model_accuracy_compact}

\subsection{Policy Training and Deployment}\label{sec:downstream_training}
After UAN training, we freeze its weights and insert it into downstream task training as a deterministic residual correction. We remove all robot-specific DR, but still randomize the terrain mesh, ground friction, and external pushes to encourage robust, deployable behaviors. The task policy outputs joint-position targets, but the simulator receives the UAN-corrected target from \eqref{eq:uan_cmd}. This exposes the downstream policy to actuation behavior closer to the hardware, so high-reward behaviors must remain feasible under the corrected actuator response. At deployment, only the learned task policy is run on the robot.

\section{Experimental Setup}\label{sec:exp_setup}

We evaluate Contact-UAN on a 29-DoF Unitree G1 humanoid. 
The nominal simulator uses BeyondMimic G1 parameters~\cite{zakka2026mjlab,liao2025beyondmimic}, including PD gains and manufacturer-provided actuator inertias. No additional robot-specific SysID is performed before Contact-UAN training. The method requires useful rigid-body and contact dynamics, but not an accurately identified actuator model.

For data collection, we deploy a WTW-style policy~\cite{margolis2023walk} on hardware for $6.5$ minutes while varying the WTW commands. 
For downstream evaluation, we compare running, force-control, and jumping policies trained in the frozen UAN-augmented and nominal simulators. No-UAN policies use additional robot-specific randomization, following BeyondMimic~\cite{liao2025beyondmimic}; UAN policies do not (Appendix~\ref{app:sec:downstream-training}). Running follows the mjlab G1 velocity-tracking task~\cite{zakka2026mjlab} (Appendix \ref{app:ssec:running}); force control adapts legged force-control objectives~\cite{portela2024learning} (Appendix \ref{app:ssec:force}); and jumping uses reference tracking~\cite{liao2025beyondmimic}, followed by task-reward optimization for height (Appendix \ref{app:ssec:jumping}). These tasks are chosen because their performance benefits from accurate actuation while the robot
exchanges forces with the environment: running benefits from efficient ground-force generation,
force control and impedance from precise external force regulation, and jumping from impulse generation
and impact absorption. 

For controlled sim-to-sim evaluation, we construct \texttt{easy}, \texttt{medium}, and \texttt{hard} target simulators by scaling deterministic perturbations to damping, friction loss, action delay, encoder bias, and armature; exact values are in Appendix~\ref{app:sec:sim2sim_setup}. Videos are provided on our \textbf{\href{https://contact-uan.csail.mit.edu/}{project website}}.

\section{Results}\label{sec:result}

\subsection{Residual World Model Accuracy} \label{sec:residual_accuracy}
\input{figures/camera_ready/torque_replay_horizontal}
\textbf{Does our estimation pipeline accurately recover foot penetration?}
We evaluate the contact-pose estimator in the \texttt{hard} sim-to-sim setting, where ground-truth contact pose is available. Fig.~\ref{fig:state_estimation}A shows that directly regressing foot height over all gait phases yields large errors, because swing-foot variation dominates the supervised objective. In contrast, our contact-conditioned estimator, which first identifies contact status and then estimates pose only for confident contacts, reduces contact-height RMSE by an order of magnitude. This shows that the proposed estimation pipeline provides sufficiently accurate contact initialization for replay-based residual learning.

\textbf{Does Contact-UAN reduce the sim-to-real gap?}
Relabeling improves replay over direct initialization, and adding the UAN further reduces 24-step joint-position MSE on WTW hardware data (Fig.~\ref{fig:state_estimation}B), with representative knee trajectories following held-out hardware running data (Appendix Figure~\ref{fig:knee_position_replay}). On held-out running data, we replay identical recorded commands from the same relabeled initial states in the nominal and UAN-augmented simulators. For commanded torques, define aggregate usage as $U(\tau)=\sum_{t,j}|\tau_{t,j}|$ and relative mismatch as $|U(\tau^{\mathrm{sim}})-U(\tau^{\mathrm{hw}})|/U(\tau^{\mathrm{hw}})$. Contact-UAN reduces this all-joint aggregate mismatch by 69.2\% (Fig.~\ref{fig:running_torque_gap}C). Figure~\ref{fig:running_torque_gap}A--B provides representative knee-torque traces. Because both simulators use identical commands and initial states, the reduced torque mismatch reflects improved actuator modeling on motions with contacts. Appendix~\ref{app:replay_visualizations} shows qualitative open-loop replay images. 

\textbf{Are the gains explained by reset-estimation errors?} 
In the hard sim-to-sim setting, we evaluate the same trained UAN, without retraining, under estimated and ground-truth initialization, with the residual enabled or disabled. Even with ground-truth initialization, the residual reduces joint-position RMSE from 0.0720 to 0.0184 rad, supporting improved actuator modeling rather than compensation for initialization error (Fig.~\ref{fig:state_estimation}C).

\subsection{Sim-to-Real Task Transfer} \label{sec:sim2real_transfer}

\textbf{Does Contact-UAN improve torque efficiency?}
We compare the baseline and UAN policy at commanded forward speeds of $1.0$, $2.0$, and $2.5~\mathrm{m\,s^{-1}}$. For each policy and command, we run five hardware trials over a fixed $30~\mathrm{ft}$ course. We compute achieved speed from the traversal time and log joint torques over the same interval to measure torque usage. The UAN policy preserves velocity tracking while reducing torque usage relative to the baseline policy (\cref{fig:hardware_results}A,B). Since both policies achieve similar speeds, the torque reduction reflects improved efficiency rather than reduced task execution. The UAN policy also transfers to a $3.0~\mathrm{m\,s^{-1}}$ command, achieving $3.13 \pm 0.06~\mathrm{m\,s^{-1}}$ on hardware (\cref{fig:main}A). 
In contrast, the baseline policy triggers motor-overheating protection and enters safe-damping mode after a few seconds of tracking this command.
\input{figures/camera_ready/hardware_figures}

\textbf{Does Contact-UAN improve force control precision?}
We evaluate end-effector force tracking using a dynamometer. For each policy and commanded force, we perform four hardware trials, increasing the commanded downward force in increments of \SI{5}{\newton}. The robot is connected to the dynamometer through an exercise band, requiring whole-body coordination rather than solely arm motion to realize the commanded force (\cref{fig:main}B). The UAN force policy reduces overall force-tracking error by $42.8 \pm 12.1\%$ relative to the baseline (Fig.~\ref{fig:hardware_results}C). The improvement is largest at higher commanded forces, where the robot must bend its legs and coordinate lower-body posture to generate the desired force, increasing sensitivity to actuator-modeling errors. At zero commanded force, the UAN policy is qualitatively \emph{compliant}: the end-effector can be pulled in different directions while the body flexibly yields rather than resisting (\cref{fig:main}C). We also compose the force-control policy with a proprioceptive end-effector estimator and PD outer loop to perform \textit{impedance position control}. The resulting controller tracks a circular trajectory using onboard sensing alone (\cref{fig:main}D and \cref{fig:impedance}), suggesting that Contact-UAN's force-control gains transfer to higher-level compliant interaction behaviors; details are in Appendix~\ref{app:ssec:force}.

\textbf{Does Contact-UAN enable robust, dynamic motions?}
We train jumping policies in two stages: reference-motion tracking~\cite{liao2025beyondmimic}, then task-reward fine-tuning for increased jump height. Both reference-tracking policies transfer to hardware, but UAN tracks better and jumps higher. After task-reward fine-tuning, the no-UAN policy fails during landing on hardware, whereas the UAN policy transfers with greater jump height and stable landings, even under pushes (\cref{fig:main}E). This behavior is consistent with the no-UAN policy exploiting actuator-model inaccuracies. The same UAN uses only walking training data. Appendix~\ref{app:ssec:jumping} and \cref{fig:jumping} provide the training stages and images of the hardware experiments.

\subsection{Sim-to-Sim Evaluation} \label{sec:sim2sim_transfer}

\textbf{Can Contact-UAN overcome larger sim-to-real gaps?}
To test whether the UAN is effective when the nominal simulator is substantially inaccurate, we evaluate transfer in sim-to-sim experiments with progressively larger deterministic actuator perturbations. The WTW UANs were trained on 5 minutes of WTW data, while the WTW+Vel UAN was trained on 5 minutes of WTW data and 5 minutes of running data. Across the \texttt{easy}, \texttt{medium}, and \texttt{hard} target simulators, the baseline policy degrades sharply as the gap increases, while policies trained with the UAN achieve expert-level tracking error and torque usage (Fig.~\ref{fig:sim2sim_results}B--C). Adding task-specific running data to the UAN training set provides little additional benefit over the WTW data alone, suggesting that this walking dataset is sufficient to learn a residual model that generalizes to running. These results indicate that the UAN can compensate for larger actuator-modeling errors than those observed on hardware.

\input{figures/camera_ready/sim2sim_figures}

\textbf{How does replay initialization affect transfer?}
We train separate UANs on \texttt{hard} target data using direct-estimator, contact-consistent, and oracle reset variants. For each UAN, we train a downstream policy and evaluate transfer to the target simulator.
Contact-consistent resets increase 20-s survival from 49.8\% to 96.7\% relative to direct-estimator resets, while component and full-oracle resets yield 96.6--98.4\% (Fig.~\ref{fig:sim2sim_results}A). These results show that contact-consistent initialization substantially improves downstream transfer, approaching the performance obtained with oracle resets. Appendix~\ref{app:reset_protocol} details the reset protocols and oracle components; Appendix~\ref{app:sensor_corruption} reports additional retraining experiments with structured sensor corruption.

\textbf{How do alternative architectures compare?} We also examine alternative choices for the learned dynamics model.
A matched simulation comparison gives similar downstream performance for position residuals, torque residuals, and supervised torque models. Position residuals avoid torque supervision and retain the deployed position-control interface; Appendix~\ref{app:actuator_baselines} reports the complete comparison. Appendix~\ref{app:sec:rwm} provides an additional data-efficiency comparison with our G1 implementation of a fully learned offline world model \cite{li2025robotic, li2025uncertainty}.

\section{Limitations}\label{sec:limitation_future}

Our current Contact-UAN does not capture coupled actuator effects or motor-temperature-dependent effects~\cite{muller2026olaf}. Coupled actuator effects could be introduced by using a single model that observes the whole body or through cross-attention. Temperature effects are harder to identify from short datasets because motor temperature evolves slowly relative to the control loop. Capturing them may also require heat modeling or additional sensors.

\section{Related Work}\label{sec:related}

\textbf{System Identification and Domain Randomization.}
System identification estimates inertial and dynamic parameters from robot data, traditionally via least-squares regression on rigid-body dynamics~\cite{khosla1985parameter,AtkesonEstimation}, with recent geometric, convex, differentiable, and exploration-based variants improving robustness and excitation~\cite{LeeGeometric,WensingGeometricObservability,SchwendemanHumanoidSysID,le2021differentiable,sobanbabu2025sampling}. Other approaches identify simulators as black boxes using trajectory-matching~\cite{bjelonic2025towards} or distribution-matching objectives~\cite{miller2025highperformancereinforcementlearningspot}. These methods remain limited by the structure of the chosen parametric model. Domain randomization instead trains over parameter distributions~\cite{tobin2017domain,tan2018simtoreal,Lee_2020,rudin2022learningwalkminutesusing,cha2025sim,muratore2022robot}, but wide randomization can yield conservative or inefficient policies~\cite{xie2021dynamics,campanaro2024learning}. Online adaptation mitigates this by inferring latent dynamics at deployment~\cite{kumar2021rmarapidmotoradaptation,ji2022concurrent,nahrendra2023dreamwaq,long2024hybridinternalmodellearning,liu2025locoformer}. Recent humanoid systems have therefore moved toward less randomization and more accurate models~\cite{liao2025beyondmimic,sleiman2026zest,bjelonic2025towards}. Our work follows this direction by replacing actuator randomization with a residual actuator correction learned from proprioceptive hardware data.

\textbf{Robotic World Models.}
Fully learned world models replace hand-designed simulators with dynamics models trained from robot data. Dyna-style approaches alternate data collection, model learning, and policy optimization through imagined rollouts~\cite{sutton1991dyna,wu2023daydreamer,hansen2025hierarchical}, while RWM and RWM-U extend this paradigm to legged robots in online~\cite{li2025robotic} and fully offline settings~\cite{li2025uncertainty}. Such models reduce the structural bias of parametric simulators but require large datasets and mechanisms to prevent exploitation outside the data distribution~\cite{hou2026world}. We instead learn a residual world model: the simulator retains rigid-body dynamics, contact, and integration, while learning is restricted to the actuator channel. This physics prior mitigates reward hacking.

\textbf{Residual Models and Actuator Networks.}
Residual learning combines analytical physics with learned corrections across object, system, simulator, and calibration levels~\cite{zeng2020tossingbot,gruenstein2021residual,sontakke2023residual,golemo2018sim,gao2024sim,jiang2021simgan,yu2025neural}. At the actuator level, supervised actuator networks predict output torques from tracking histories and have enabled effective sim-to-real transfer~\cite{Hwangbo_2019}, but require torque sensing; current-based alternatives reduce sensing requirements but cannot capture all gearbox effects~\cite{SchwendemanHumanoidSysID}. UAN removes torque supervision by learning residual actions that match joint transitions in simulation with recorded hardware data~\cite{fey2025bridging}, but was previously demonstrated only on a manipulator arm mounted on a quadruped. Closely related humanoid methods such as ASAP and MOSAIC also learn residual corrections, but rely on motion capture or target interface-specific gaps rather than reusable actuator dynamics~\cite{he2025asap,sun2026mosaic}. We extend UAN to a floating-base 29-DoF humanoid by using proprioceptive state estimators for base velocity and contact pose, enabling contact-aligned residual model learning without external sensing.

\section{Conclusion}\label{sec:conclusion}
We present \emph{Contact-UAN}, a residual world-modeling approach for humanoid control that preserves the structure of an analytical simulator while learning actuator-level corrections from offline proprioceptive data. To enable replay-based learning from humanoid motion with contacts, we estimate base velocity, contact status, and simulator-side contact pose from onboard observations, allowing hardware trajectories to be initialized in simulation with consistent contact forces. Our UAN-augmented simulator improves transfer of whole-body policies across diverse tasks using only 6.5 minutes of real-world walking data: 20\% lower running torque usage, 43\% lower force-tracking error, compliant interaction, accurate impedance control, and higher jumps with stable landings after task-reward optimization. Reset ablations support contact-consistent initialization, and the residual's benefit persists under ground-truth initialization. Sim-to-sim experiments further show that Contact-UAN remains effective under larger controlled actuator gaps. These results suggest that residual world models provide a data-efficient alternative to fully learned dynamics models, retaining the generalization benefits of physics simulation while improving fidelity where the sim-to-real gap is most consequential.

\clearpage
\acknowledgments{We thank the members of the Improbable AI lab, especially Venkateshkrishna Venganallore Parsuram, Michelle Wang, and Ryan Bahlous-Boldi for supporting hardware experiments. We also thank Patrick Wensing and members of the Harvard Biorobotics Lab for helpful discussions. The research was partly supported by the Ministry of Trade, Industry, and Energy (MOTIE), Korea, under the ``Global Industrial Technology Cooperation Center program" supervised by the Korea Institute for Advancement of Technology (KIAT) under Grant No. P0028435 and the National Science Foundation Graduate Research Fellowship under Grant No. 2141064. This work was also partly supported by the Sony Research Award.}

\bibliography{corl_2026_template_submission/reference}  

\clearpage

\appendix

\input{appendices/uan_method_appendices_new_new}

\end{document}

%% file: figures/camera_ready/model_accuracy_compact.tex
\begin{figure}[tbp]
\centering
\input{figures/camera_ready/model_accuracy_compact_plot}
\vspace{-1em}
\caption{\textbf{Contact estimation and replay accuracy.} (A) Contact-height RMSE for estimators trained on both stance and swing (Direct) or stance only (Stance), evaluated in the \texttt{hard} target simulator. (B) 24-step hardware joint-position MSE on WTW training data. (C) Sim-to-sim 24-step joint-position RMSE. The same UAN substantially reduces replay error under both estimated and ground-truth initialization, supporting actuator-model correction beyond compensation for state-estimation errors.}
\label{fig:state_estimation}
\vspace{-1em}
\end{figure}
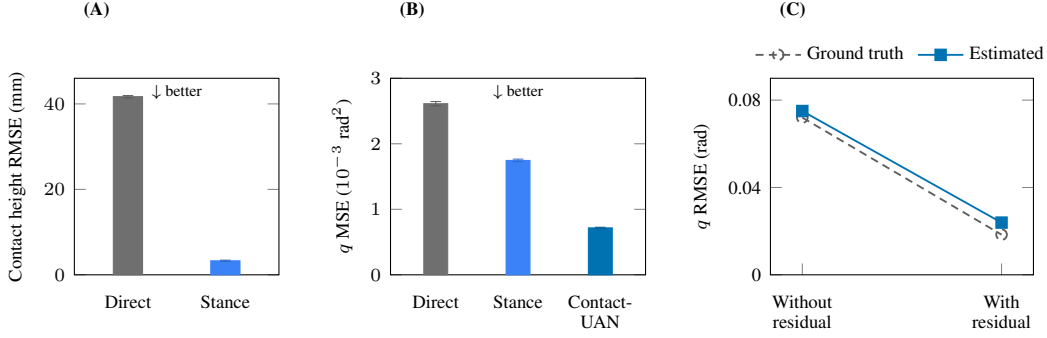

%% file: figures/camera_ready/model_accuracy_compact_plot.tex
\definecolor{OursBlue}{HTML}{3B82F6}
\begin{tikzpicture}
\begin{axis}[
                scale only axis,
                at={(29pt,0pt)},anchor=south west,
                width=77pt,height=2.6cm,
                ybar,
                bar width=11pt,
                ymin=0,
                ymax=46,
                xmin=0.45,
                xmax=2.55,
                xtick={1,2},
                xticklabels={Direct,Stance},
                xtick pos=bottom,
                xtick style={draw=none},
                tick align=inside,
                filter discard warning=false,
                clip=false,
                ylabel={Contact height RMSE (mm)},
                tick label style={font=\scriptsize},
                label style={font=\scriptsize},
                xticklabel style={font=\scriptsize, align=center,yshift=-4pt},
            ]
                \addplot+[
                    fill=BaselineGray,
                    draw=BaselineGray,
                    bar shift=0pt,
                    restrict x to domain=0.5:1.5,
                    unbounded coords=discard,
                    error bars/.cd,
                    y dir=both,
                    y explicit,
                    error mark=-,
                    error bar style={draw=BaselineGray!80!black, line width=0.35pt},
                    error mark options={draw=BaselineGray!80!black, line width=0.35pt, rotate=90, mark size=2.0pt},
                ] table[
                    x=bar_x,
                    y=value,
                    y error minus=err_low,
                    y error plus=err_high,
                    col sep=comma
                ] {figures/data/state_estimation_panel_a.csv};
                \addplot+[
                    fill=OursBlue,
                    draw=OursBlue,
                    bar shift=0pt,
                    restrict x to domain=1.5:2.5,
                    unbounded coords=discard,
                    error bars/.cd,
                    y dir=both,
                    y explicit,
                    error mark=-,
                    error bar style={draw=OursBlue!80!black, line width=0.35pt},
                    error mark options={draw=OursBlue!80!black, line width=0.35pt, rotate=90, mark size=2.0pt},
                ] table[
                    x=bar_x,
                    y=value,
                    y error minus=err_low,
                    y error plus=err_high,
                    col sep=comma
                ] {figures/data/state_estimation_panel_a.csv};
                \node[anchor=south west, font=\scriptsize] at (rel axis cs:0.0,1.26) {\textbf{(A)}};
                \node[
                    anchor=north,
                    font=\tiny,
                    inner sep=1pt,
                    fill=white,
                    fill opacity=0.75,
                    text opacity=1
                ] at (rel axis cs:0.5,0.98) {$\downarrow$ better};
            \end{axis}
\begin{axis}[
                scale only axis,
                at={(148pt,0pt)},anchor=south west,
                width=96pt,height=2.6cm,
                ybar,
                bar width=9pt,
                ymin=0,
                ymax=3.0,
                xmin=0.45,
                xmax=3.55,
                xtick={1,2,3},
                xticklabels={Direct,Stance,{\shortstack{Contact-\\UAN}}},
                xtick pos=bottom,
                xtick style={draw=none},
                tick align=inside,
                filter discard warning=false,
                clip=false,
                ylabel={$q$ MSE ($10^{-3}$ rad$^2$)},
                tick label style={font=\scriptsize},
                label style={font=\scriptsize},
                xticklabel style={font=\scriptsize, align=center,yshift=-4pt},
            ]
                \addplot+[
                    fill=BaselineGray,
                    draw=BaselineGray,
                    bar shift=0pt,
                    restrict x to domain=0.5:1.5,
                    unbounded coords=discard,
                    error bars/.cd,
                    y dir=both,
                    y explicit,
                    error mark=-,
                    error bar style={draw=BaselineGray!80!black, line width=0.35pt},
                    error mark options={draw=BaselineGray!80!black, line width=0.35pt, rotate=90, mark size=2.0pt},
                ] table[
                    x=bar_x,
                    y=value,
                    y error minus=err_low,
                    y error plus=err_high,
                    col sep=comma
                ] {figures/data/state_estimation_panel_b.csv};
                \addplot+[
                    fill=OursBlue,
                    draw=OursBlue,
                    bar shift=0pt,
                    restrict x to domain=1.5:2.5,
                    unbounded coords=discard,
                    error bars/.cd,
                    y dir=both,
                    y explicit,
                    error mark=-,
                    error bar style={draw=OursBlue!80!black, line width=0.35pt},
                    error mark options={draw=OursBlue!80!black, line width=0.35pt, rotate=90, mark size=2.0pt},
                ] table[
                    x=bar_x,
                    y=value,
                    y error minus=err_low,
                    y error plus=err_high,
                    col sep=comma
                ] {figures/data/state_estimation_panel_b.csv};
                \addplot+[
                    fill=UANTeal,
                    draw=UANTeal,
                    bar shift=0pt,
                    restrict x to domain=2.5:3.5,
                    unbounded coords=discard,
                    error bars/.cd,
                    y dir=both,
                    y explicit,
                    error mark=-,
                    error bar style={draw=UANTeal!80!black, line width=0.35pt},
                    error mark options={draw=UANTeal!80!black, line width=0.35pt, rotate=90, mark size=2.0pt},
                ] table[
                    x=bar_x,
                    y=value,
                    y error minus=err_low,
                    y error plus=err_high,
                    col sep=comma
                ] {figures/data/state_estimation_panel_b.csv};
                \node[anchor=south west, font=\scriptsize] at (rel axis cs:0.0,1.26) {\textbf{(B)}};
                \node[
                    anchor=north,
                    font=\tiny,
                    inner sep=1pt,
                    fill=white,
                    fill opacity=0.75,
                    text opacity=1
                ] at (rel axis cs:0.5,0.98) {$\downarrow$ better};
            \end{axis}
\begin{axis}[
scale only axis,
at={(291pt,0pt)},anchor=south west,
width=99pt,height=2.6cm,
xmin=-0.16,xmax=1.16,ymin=0,ymax=0.09,
xtick={0,1},xticklabels={{\shortstack{Without\\residual}},{\shortstack{With\\residual}}},
xticklabel style={align=center,yshift=-4pt},
ytick={0,0.04,0.08},yticklabels={0,0.04,0.08},
ylabel={$q$ RMSE (rad)},
tick label style={font=\scriptsize},label style={font=\scriptsize},
tick align=inside,xtick pos=bottom,scaled y ticks=false,
legend style={font=\scriptsize,draw=none,at={(0.5,1.02)},anchor=south,legend columns=2,/tikz/every even column/.append style={column sep=3pt}},
clip=false]
\addplot[BaselineGray!85!black,densely dashed,mark=o,mark size=2pt,line width=0.7pt] coordinates {(0,0.0720) (1,0.0184)};
\addlegendentry{Ground truth}
\addplot[UANTeal,mark=square*,mark size=2pt,line width=0.7pt] coordinates {(0,0.0750) (1,0.0239)};
\addlegendentry{Estimated}
\node[anchor=south west,font=\scriptsize] at (rel axis cs:0,1.26) {\textbf{(C)}};
\end{axis}
\end{tikzpicture}

%% file: figures/camera_ready/torque_replay_horizontal.tex
\begin{figure}[!t]
\centering
\pgfplotstableread[col sep=comma]{figures/data/running_knee_torque_trace_representative.csv}\horizontalrunningtrace
\pgfplotstableread[col sep=comma]{figures/data/running_torque_gap_bars.csv}\horizontalrunningbars
\pgfplotstablegetelem{0}{net_torque_error_mean}\of\horizontalrunningbars
\edef\HorizontalNominalMean{\pgfplotsretval}
\pgfplotstablegetelem{0}{net_torque_error_sem}\of\horizontalrunningbars
\edef\HorizontalNominalSem{\pgfplotsretval}
\pgfplotstablegetelem{1}{net_torque_error_mean}\of\horizontalrunningbars
\edef\HorizontalUanMean{\pgfplotsretval}
\pgfplotstablegetelem{1}{net_torque_error_sem}\of\horizontalrunningbars
\edef\HorizontalUanSem{\pgfplotsretval}
\pgfmathsetmacro{\HorizontalTorqueReduction}{100*(\HorizontalNominalMean-\HorizontalUanMean)/\HorizontalNominalMean}
\pgfplotsset{horizontalrunningtrace/.style={
    width=\linewidth,height=3.45cm,
    xmin=0,xmax=0.46,
    grid=major,grid style={line width=0.2pt,draw=gray!20},
    tick label style={font=\scriptsize},label style={font=\scriptsize},
    xlabel={Time (s)},ylabel={Torque (N\,m)},xtick={0,0.2,0.4},
    axis x line*=bottom,axis y line*=left,tick align=inside,
    tick style={line width=0.25pt},axis line style={line width=0.3pt},
    title style={font=\scriptsize},
}}
{\scriptsize
\begin{tabular}{@{}ccc@{}}
\tikz{\draw[black,line width=0.75pt] (0,0.05)--(0.24,0.05);} Hardware &
\tikz{\draw[UANTeal,line width=0.75pt,densely dashed] (0,0.05)--(0.24,0.05);} UAN &
\tikz{\draw[BaselineGray!85!black,line width=0.75pt,densely dashed] (0,0.05)--(0.24,0.05);} No UAN
\end{tabular}\par}
\vspace{0.3em}
\begin{minipage}[t]{0.315\linewidth}
\vspace{0pt}\centering
\begin{tikzpicture}
\begin{axis}[horizontalrunningtrace,ymin=-25,ymax=-5,ytick={-25,-15,-5},
    title={\textbf{(A)} Left knee}]
\addplot[black,line width=0.8pt] table[x=time,y=left_knee_hw]{\horizontalrunningtrace};
\addplot[UANTeal,line width=0.8pt,densely dashed] table[x=time,y=left_knee_uan]{\horizontalrunningtrace};
\addplot[BaselineGray!85!black,line width=0.8pt,densely dashed] table[x=time,y=left_knee_nominal]{\horizontalrunningtrace};
\end{axis}
\end{tikzpicture}
\end{minipage}\hfill
\begin{minipage}[t]{0.315\linewidth}
\vspace{0pt}\centering
\begin{tikzpicture}
\begin{axis}[horizontalrunningtrace,ymin=-40,ymax=-10,ytick={-40,-25,-10},
    title={\textbf{(B)} Right knee}]
\addplot[black,line width=0.8pt] table[x=time,y=right_knee_hw]{\horizontalrunningtrace};
\addplot[UANTeal,line width=0.8pt,densely dashed] table[x=time,y=right_knee_uan]{\horizontalrunningtrace};
\addplot[BaselineGray!85!black,line width=0.8pt,densely dashed] table[x=time,y=right_knee_nominal]{\horizontalrunningtrace};
\end{axis}
\end{tikzpicture}
\end{minipage}\hfill
\begin{minipage}[t]{0.315\linewidth}
\vspace{0pt}\centering
\begin{tikzpicture}
\begin{axis}[
    width=\linewidth,height=3.45cm,ybar,bar width=13pt,
    ymin=0,ymax=0.21,xmin=-0.65,xmax=1.65,
    xtick={0,1},xticklabels={No UAN,UAN},ytick={0,0.1,0.2},
    yticklabels={0,0.10,0.20},ylabel={Torque-usage relative error},
    xtick pos=bottom,xtick style={draw=none},tick align=inside,clip=false,
    tick label style={font=\scriptsize},label style={font=\scriptsize},
    xticklabel style={font=\scriptsize,align=center},
    title={\textbf{(C)} Aggregate mismatch},title style={font=\scriptsize},
]
\addplot+[
    fill=BaselineGray,draw=BaselineGray,bar shift=0pt,
    error bars/.cd,y dir=both,y explicit,error mark=-,
    error bar style={draw=BaselineGray!80!black,line width=0.35pt},
    error mark options={draw=BaselineGray!80!black,line width=0.35pt,rotate=90,mark size=2pt}
] coordinates {(0,\HorizontalNominalMean) +- (0,\HorizontalNominalSem)};
\addplot+[
    fill=UANTeal,draw=UANTeal,bar shift=0pt,
    error bars/.cd,y dir=both,y explicit,error mark=-,
    error bar style={draw=UANTeal!80!black,line width=0.35pt},
    error mark options={draw=UANTeal!80!black,line width=0.35pt,rotate=90,mark size=2pt}
] coordinates {(1,\HorizontalUanMean) +- (0,\HorizontalUanSem)};
\node[anchor=north,font=\scriptsize\bfseries,fill=white,inner sep=1pt,align=center]
    at (rel axis cs:0.71,0.66) {\pgfmathprintnumber[fixed,precision=1]{\HorizontalTorqueReduction}\%\\reduction};
\end{axis}
\end{tikzpicture}
\end{minipage}
\caption{\textbf{Held-out running torque replay.} (A, B) Representative left- and right-knee commanded-torque traces under identical relabeled initial states. (C) Aggregate commanded torque-usage relative error decreases by $69.2\%$ with UAN.}
\label{fig:running_torque_gap}
\vspace{-1em}
\end{figure}
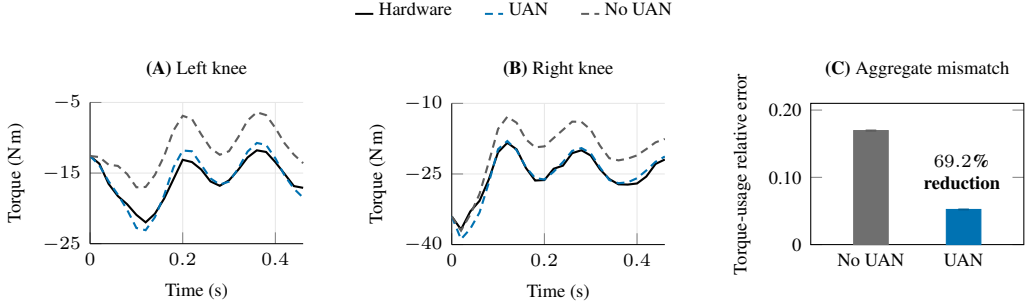

%% file: figures/camera_ready/hardware_figures.tex
\begin{figure}[tbp]
    \centering
    \begin{minipage}[t]{0.64\linewidth}
        \centering
        \scriptsize
        \begin{tabular}{@{}cc@{}}
            \tikz{\draw[BaselineGray!85!black, line width=0.7pt] (0,0.06) -- (0.24,0.06);
                  \draw[fill=BaselineGray!85!black, draw=BaselineGray!85!black] (0.12,0.06) rectangle (0.16,0.10);}
            Baseline velocity &
            \tikz{\draw[UANTeal, line width=0.7pt] (0,0.06) -- (0.24,0.06);
                  \draw[fill=UANTeal, draw=UANTeal] (0.12,0.08) circle (0.035);}
            WTW-UAN-aug velocity
        \end{tabular}
    \end{minipage}
    \hfill
    \begin{minipage}[t]{0.35\linewidth}
        \centering
        \scriptsize
        \begin{tabular}{@{}cc@{}}
            \tikz{\draw[BaselineGray!85!black, line width=0.7pt] (0,0.06) -- (0.24,0.06);
                  \draw[fill=BaselineGray!85!black, draw=BaselineGray!85!black] (0.12,0.06) rectangle (0.16,0.10);}
            Baseline force &
            \tikz{\draw[UANTeal, line width=0.7pt] (0,0.06) -- (0.24,0.06);
                  \draw[fill=UANTeal, draw=UANTeal] (0.12,0.08) circle (0.035);}
            WTW-UAN-aug force
        \end{tabular}
    \end{minipage}
    \vspace{-0.45em}

    \begin{minipage}[t]{0.32\linewidth}
        \centering
        \begin{tikzpicture}
            \begin{axis}[
                width=\linewidth,
                height=4.0cm,
                ymin=0,
                ymax=3.35,
                xmin=0.8,
                xmax=3.15,
                xtick={1,2,2.5,3},
                xticklabels={1.0,2.0,2.5,3.0},
                xlabel={Command (m/s)},
                ylabel={Achieved speed (m/s)},
                tick label style={font=\scriptsize},
                label style={font=\scriptsize},
            ]
                \addplot[
                    black,
                    dashed,
                    line width=0.7pt,
                ] coordinates {(1,1) (3,3)};
                \addplot[
                    BaselineGray!85!black,
                    line width=0.7pt,
                ] table[
                    x=baseline_speed_mps,
                    y=baseline_speed_fit_mps,
                    col sep=comma
                ] {figures/data/velocity_speed_fit_lines_hardware.csv};
                \addplot[
                    UANTeal,
                    line width=0.7pt,
                ] table[
                    x=uan_speed_mps,
                    y=uan_speed_fit_mps,
                    col sep=comma
                ] {figures/data/velocity_speed_fit_lines_hardware.csv};
                \addplot+[
                    only marks,
                    BaselineGray!85!black,
                    mark=square*,
                    mark size=1.35pt,
                    forget plot,
                    error bars/.cd,
                    y dir=both,
                    y explicit,
                    error mark=-,
                    error bar style={draw=BaselineGray!80!black, line width=0.35pt},
                    error mark options={draw=BaselineGray!80!black, line width=0.35pt, rotate=90, mark size=2.0pt},
                ] table[
                    x=speed_mps,
                    y=baseline_speed_mean_mps,
                    y error=baseline_speed_std_mps,
                    col sep=comma
                ] {figures/data/velocity_relative_summary.csv};
                \addplot+[
                    only marks,
                    UANTeal,
                    mark=*,
                    mark size=1.35pt,
                    forget plot,
                    error bars/.cd,
                    y dir=both,
                    y explicit,
                    error mark=-,
                    error bar style={draw=UANTeal!80!black, line width=0.35pt},
                    error mark options={draw=UANTeal!80!black, line width=0.35pt, rotate=90, mark size=2.0pt},
                ] table[
                    x=speed_mps,
                    y=uan_speed_mean_mps,
                    y error=uan_speed_std_mps,
                    col sep=comma
                ] {figures/data/velocity_relative_summary.csv};
                \addplot+[
                    only marks,
                    UANTeal,
                    mark=*,
                    mark size=1.35pt,
                    forget plot,
                    restrict x to domain=2.75:3.25,
                    error bars/.cd,
                    y dir=both,
                    y explicit,
                    error mark=-,
                    error bar style={draw=UANTeal!80!black, line width=0.35pt},
                    error mark options={draw=UANTeal!80!black, line width=0.35pt, rotate=90, mark size=2.0pt},
                ] table[
                    x=speed_mps,
                    y=uan_speed_mean_mps,
                    y error=uan_speed_std_mps,
                    col sep=comma
                ] {figures/data/velocity_tracking_error.csv};
                \draw[
                    BaselineGray!85!black,
                    densely dashed,
                    line width=0.55pt
                ] (axis cs:2.5,2.9045) -- (axis cs:3,0.05);
                \addplot+[
                    only marks,
                    BaselineGray!85!black,
                    mark=x,
                    mark size=2.3pt,
                    line width=0.7pt,
                    forget plot,
                ] coordinates {(3,0.05)};
                \node[
                    anchor=south east,
                    font=\tiny,
                    align=center,
                    inner sep=1pt,
                    fill=white,
                    fill opacity=0.85,
                    text opacity=1
                ] at (axis cs:2.96,0.28) {safe damping};
                \node[anchor=north west, font=\scriptsize] at (rel axis cs:0.02,0.98) {\textbf{(A)}};
            \end{axis}
        \end{tikzpicture}
    \end{minipage}
    \hfill
    \begin{minipage}[t]{0.32\linewidth}
        \centering
        \begin{tikzpicture}
            \begin{axis}[
                width=\linewidth,
                height=4.0cm,
                ybar,
                bar width=8pt,
                ymin=-7,
                ymax=36,
                xmin=0.75,
                xmax=2.75,
                xtick={1,2,2.5},
                xticklabels={1.0,2.0,2.5},
                xtick pos=bottom,
                tick align=inside,
                xlabel={Command (m/s)},
                ylabel={Torque reduction (\%)},
                tick label style={font=\scriptsize},
                label style={font=\scriptsize},
            ]
                \addplot+[
                    fill=UANTeal,
                    draw=UANTeal,
                    bar shift=0pt,
                    error bars/.cd,
                    y dir=both,
                    y explicit,
                    error mark=-,
                    error bar style={draw=UANTeal!80!black, line width=0.35pt},
                    error mark options={draw=UANTeal!80!black, line width=0.35pt, rotate=90, mark size=2.0pt},
                ] table[
                    x=speed_mps,
                    y=tau2_reduction_percent,
                    y error=tau2_reduction_std_percent,
                    col sep=comma
                ] {figures/data/velocity_relative_summary.csv};
                \draw[black, thin, dashed] (axis cs:0.75,0) -- (axis cs:2.75,0);
                \node[anchor=north west, font=\scriptsize] at (rel axis cs:0.02,0.98) {\textbf{(B)}};
                \node[
                    anchor=north,
                    font=\tiny,
                    inner sep=1pt,
                    fill=white,
                    fill opacity=0.75,
                    text opacity=1
                ] at (rel axis cs:0.5,0.98) {$\uparrow$ better};
            \end{axis}
        \end{tikzpicture}
    \end{minipage}
    \hfill
    \begin{minipage}[t]{0.32\linewidth}
        \centering
        \begin{tikzpicture}
            \begin{axis}[
                width=\linewidth,
                height=4.0cm,
                xmin=0,
                xmax=22,
                ymin=0,
                ymax=22,
                xtick={0,5,10,15,20},
                ytick={0,5,10,15,20},
                xtick pos=bottom,
                xlabel={Commanded force (N)},
                ylabel={Measured force (N)},
                tick label style={font=\scriptsize},
                label style={font=\scriptsize},
            ]
                \addplot[
                    black,
                    dashed,
                    line width=0.7pt,
                ] coordinates {(0,0) (22,22)};
                \addplot[
                    BaselineGray!85!black,
                    line width=0.7pt,
                ] table[
                    x=force_cmd_n,
                    y=baseline_force_fit_n,
                    col sep=comma
                ] {figures/data/force_control_fit_lines.csv};
                \addplot[
                    UANTeal,
                    line width=0.7pt,
                ] table[
                    x=force_cmd_n,
                    y=uan_force_fit_n,
                    col sep=comma
                ] {figures/data/force_control_fit_lines.csv};
                \addplot+[
                    only marks,
                    BaselineGray!85!black,
                    mark=square*,
                    mark size=1.35pt,
                    forget plot,
                    restrict x to domain=0:21,
                    unbounded coords=discard,
                    error bars/.cd,
                    y dir=both,
                    y explicit,
                    error mark=-,
                    error bar style={draw=BaselineGray!80!black, line width=0.35pt},
                    error mark options={draw=BaselineGray!80!black, line width=0.35pt, rotate=90, mark size=2.0pt},
                ] table[
                    x=force_cmd_n,
                    y=baseline_force_mean_n,
                    y error=baseline_force_std_n,
                    col sep=comma
                ] {figures/data/force_control_comparison.csv};
                \addplot+[
                    only marks,
                    UANTeal,
                    mark=*,
                    mark size=1.35pt,
                    forget plot,
                    restrict x to domain=0:21,
                    unbounded coords=discard,
                    error bars/.cd,
                    y dir=both,
                    y explicit,
                    error mark=-,
                    error bar style={draw=UANTeal!80!black, line width=0.35pt},
                    error mark options={draw=UANTeal!80!black, line width=0.35pt, rotate=90, mark size=2.0pt},
                ] table[
                    x=force_cmd_n,
                    y=uan_force_mean_n,
                    y error=uan_force_std_n,
                    col sep=comma
                ] {figures/data/force_control_comparison.csv};
                \node[anchor=north west, font=\scriptsize] at (rel axis cs:0.02,0.98) {\textbf{(C)}};
            \end{axis}
        \end{tikzpicture}
    \end{minipage}
    \caption{\textbf{Improved running and force control sim-to-real transfer.} UAN running policy retains tracking ability (A) while improving efficiency (B). The baseline marker at $3.0~\mathrm{m\,s^{-1}}$ denotes protection-triggered safe damping. UAN force policy reduced average force tracking error by $42.8 \pm 12.1 \%$ (C). 
    }
    \label{fig:hardware_results}
    \vspace{-1em}
\end{figure}
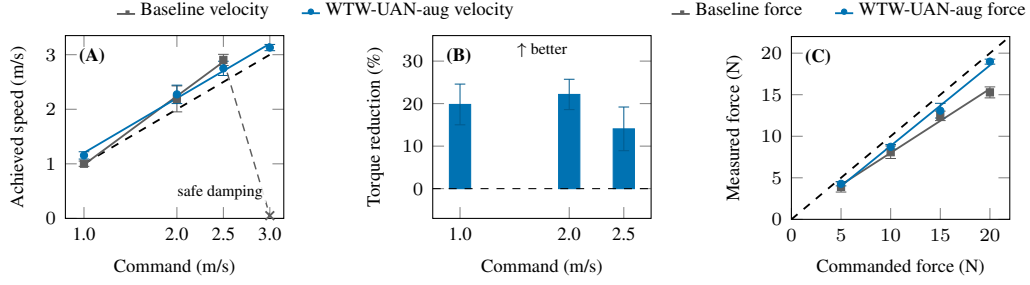

%% file: figures/camera_ready/sim2sim_figures.tex
\definecolor{WTWGreen}{HTML}{009E73}
\definecolor{ExpertOrange}{HTML}{D55E00}
\begin{figure}[tbp]
\centering
\hfill\begin{minipage}{0.66\linewidth}
\centering\scriptsize
\begin{tabular}{@{}cccc@{}}
\tikz{\draw[fill=BaselineGray,draw=BaselineGray] (0,0) rectangle (0.12,0.12);} Baseline &
\tikz{\draw[fill=UANTeal,draw=UANTeal] (0,0) rectangle (0.12,0.12);} WTW UAN &
\tikz{\draw[fill=WTWGreen,draw=WTWGreen] (0,0) rectangle (0.12,0.12);} WTW+Vel UAN &
\tikz{\draw[fill=ExpertOrange,draw=ExpertOrange] (0,0) rectangle (0.12,0.12);} Expert
\end{tabular}
\end{minipage}\par
\vspace{0.3em}
\begin{minipage}[t]{0.32\linewidth}
\vspace{0pt}\centering
{\scriptsize\textbf{(A)} Reset quality and transfer}\par
\vspace{0.35cm}
\includegraphics[width=\linewidth]{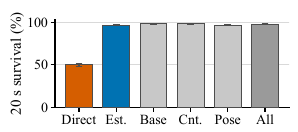}
\end{minipage}\hfill
\begin{minipage}[t]{0.32\linewidth}
\vspace{0pt}\centering
{\scriptsize\textbf{(B)} Velocity tracking}\par
\vspace{0.1cm}
\begin{tikzpicture}
\begin{axis}[
                    width=\linewidth,
                    height=3.45cm,
                    ybar,
                    bar width=4pt,
                    ymin=0,
                    ymax=0.95,
                    ytick={0,0.25,0.5,0.75},
                    enlarge x limits=0.24,
                    xtick={1,2,3},
                    xticklabels={Easy,Medium,Hard},
                    xtick pos=bottom,
                    ylabel={XY RMSE (m/s)},
                    tick label style={font=\scriptsize},
                    label style={font=\scriptsize},
                ]
                    \addplot+[
                        fill=BaselineGray,
                        draw=BaselineGray,
                        bar shift=-7.2pt,
                        error bars/.cd,
                        y dir=both,
                        y explicit,
                        error mark=-,
                        error bar style={draw=BaselineGray!80!black, line width=0.35pt},
                        error mark options={draw=BaselineGray!80!black, line width=0.35pt, rotate=90, mark size=2.0pt},
                    ] table[
                        x=profile_x,
                        y=baseline_raw_rmse_mean_mps,
                        y error minus=baseline_raw_rmse_ci95_minus_mps,
                        y error plus=baseline_raw_rmse_ci95_plus_mps,
                        col sep=comma
                    ] {figures/data/sim2sim_velocity_metrics.csv};

                    \addplot+[
                        fill=UANTeal,
                        draw=UANTeal,
                        bar shift=-2.4pt,
                        error bars/.cd,
                        y dir=both,
                        y explicit,
                        error mark=-,
                        error bar style={draw=UANTeal!80!black, line width=0.35pt},
                        error mark options={draw=UANTeal!80!black, line width=0.35pt, rotate=90, mark size=2.0pt},
                    ] table[
                        x=profile_x,
                        y=wtw_uan_raw_rmse_mean_mps,
                        y error minus=wtw_uan_raw_rmse_ci95_minus_mps,
                        y error plus=wtw_uan_raw_rmse_ci95_plus_mps,
                        col sep=comma
                    ] {figures/data/sim2sim_velocity_metrics.csv};

                    \addplot+[
                        fill=WTWGreen,
                        draw=WTWGreen,
                        bar shift=2.4pt,
                        error bars/.cd,
                        y dir=both,
                        y explicit,
                        error mark=-,
                        error bar style={draw=WTWGreen!80!black, line width=0.35pt},
                        error mark options={draw=WTWGreen!80!black, line width=0.35pt, rotate=90, mark size=2.0pt},
                    ] table[
                        x=profile_x,
                        y=wtw_vel_uan_raw_rmse_mean_mps,
                        y error minus=wtw_vel_uan_raw_rmse_ci95_minus_mps,
                        y error plus=wtw_vel_uan_raw_rmse_ci95_plus_mps,
                        col sep=comma
                    ] {figures/data/sim2sim_velocity_metrics.csv};

                    \addplot+[
                        fill=ExpertOrange,
                        draw=ExpertOrange,
                        bar shift=7.2pt,
                        error bars/.cd,
                        y dir=both,
                        y explicit,
                        error mark=-,
                        error bar style={draw=ExpertOrange!80!black, line width=0.35pt},
                        error mark options={draw=ExpertOrange!80!black, line width=0.35pt, rotate=90, mark size=2.0pt},
                    ] table[
                        x=profile_x,
                        y=expert_raw_rmse_mean_mps,
                        y error minus=expert_raw_rmse_ci95_minus_mps,
                        y error plus=expert_raw_rmse_ci95_plus_mps,
                        col sep=comma
                    ] {figures/data/sim2sim_velocity_metrics.csv};

                    \node[anchor=north, font=\tiny, inner sep=1pt,
                        fill=white, fill opacity=0.75, text opacity=1]
                        at (rel axis cs:0.5,0.98) {$\downarrow$ better};
                \end{axis}
\end{tikzpicture}
\end{minipage}\hfill
\begin{minipage}[t]{0.32\linewidth}
\vspace{0pt}\centering
{\scriptsize\textbf{(C)} Torque usage}\par
\vspace{0.1cm}
\begin{tikzpicture}
\begin{axis}[
                    width=\linewidth,
                    height=3.45cm,
                    ybar,
                    bar width=4.0pt,
                    ymin=0,
                    ymax=3600,
                    ytick={0,1000,2000,3000},
                    enlarge x limits=0.24,
                    xtick={1,2,3},
                    xticklabels={Easy,Medium,Hard},
                    xtick pos=bottom,
                    ylabel={Torque$^2$ (N$^2$m$^2$)},
                    tick label style={font=\scriptsize},
                    label style={font=\scriptsize},
                ]
                    \addplot+[
                        fill=BaselineGray,
                        draw=BaselineGray,
                        bar shift=-7.2pt,
                        error bars/.cd,
                        y dir=both,
                        y explicit,
                        error mark=-,
                        error bar style={draw=BaselineGray!80!black, line width=0.35pt},
                        error mark options={draw=BaselineGray!80!black, line width=0.35pt, rotate=90, mark size=2.0pt},
                    ] table[
                        x=profile_x,
                        y=baseline_torque2_mean_nm2,
                        y error minus=baseline_torque2_ci95_minus_nm2,
                        y error plus=baseline_torque2_ci95_plus_nm2,
                        col sep=comma
                    ] {figures/data/sim2sim_velocity_metrics.csv};

                    \addplot+[
                        fill=UANTeal,
                        draw=UANTeal,
                        bar shift=-2.4pt,
                        error bars/.cd,
                        y dir=both,
                        y explicit,
                        error mark=-,
                        error bar style={draw=UANTeal!80!black, line width=0.35pt},
                        error mark options={draw=UANTeal!80!black, line width=0.35pt, rotate=90, mark size=2.0pt},
                    ] table[
                        x=profile_x,
                        y=wtw_uan_torque2_mean_nm2,
                        y error minus=wtw_uan_torque2_ci95_minus_nm2,
                        y error plus=wtw_uan_torque2_ci95_plus_nm2,
                        col sep=comma
                    ] {figures/data/sim2sim_velocity_metrics.csv};

                    \addplot+[
                        fill=WTWGreen,
                        draw=WTWGreen,
                        bar shift=2.4pt,
                        error bars/.cd,
                        y dir=both,
                        y explicit,
                        error mark=-,
                        error bar style={draw=WTWGreen!80!black, line width=0.35pt},
                        error mark options={draw=WTWGreen!80!black, line width=0.35pt, rotate=90, mark size=2.0pt},
                    ] table[
                        x=profile_x,
                        y=wtw_vel_uan_torque2_mean_nm2,
                        y error minus=wtw_vel_uan_torque2_ci95_minus_nm2,
                        y error plus=wtw_vel_uan_torque2_ci95_plus_nm2,
                        col sep=comma
                    ] {figures/data/sim2sim_velocity_metrics.csv};

                    \addplot+[
                        fill=ExpertOrange,
                        draw=ExpertOrange,
                        bar shift=7.2pt,
                        error bars/.cd,
                        y dir=both,
                        y explicit,
                        error mark=-,
                        error bar style={draw=ExpertOrange!80!black, line width=0.35pt},
                        error mark options={draw=ExpertOrange!80!black, line width=0.35pt, rotate=90, mark size=2.0pt},
                    ] table[
                        x=profile_x,
                        y=expert_torque2_mean_nm2,
                        y error minus=expert_torque2_ci95_minus_nm2,
                        y error plus=expert_torque2_ci95_plus_nm2,
                        col sep=comma
                    ] {figures/data/sim2sim_velocity_metrics.csv};

                    \node[anchor=north, font=\tiny, inner sep=1pt,
                        fill=white, fill opacity=0.75, text opacity=1]
                        at (rel axis cs:0.5,0.98) {$\downarrow$ better};
                \end{axis}
\end{tikzpicture}
\end{minipage}
\caption{\textbf{Sim-to-sim reset ablation and policy transfer.} (A) Reported 20-s survival after retraining UANs and policies with direct, estimated (Est.), component-oracle, or full-oracle (All) resets. Base, Cnt., and Pose replace base velocity, contact state, and stance pose, respectively, with ground truth. (B, C) Velocity tracking and squared torque across target gaps. WTW-trained UAN policies approach expert performance, with little additional benefit from task-specific running data.}
\label{fig:sim2sim_results}
\label{fig:reset_quality}
\vspace{-1em}
\end{figure}
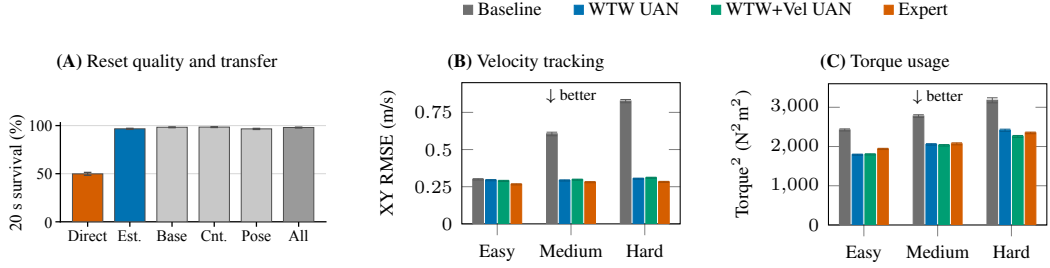

%% file: appendices/uan_method_appendices_new_new.tex
\begin{figure}
    \centering
    \includegraphics[width=0.95\linewidth]{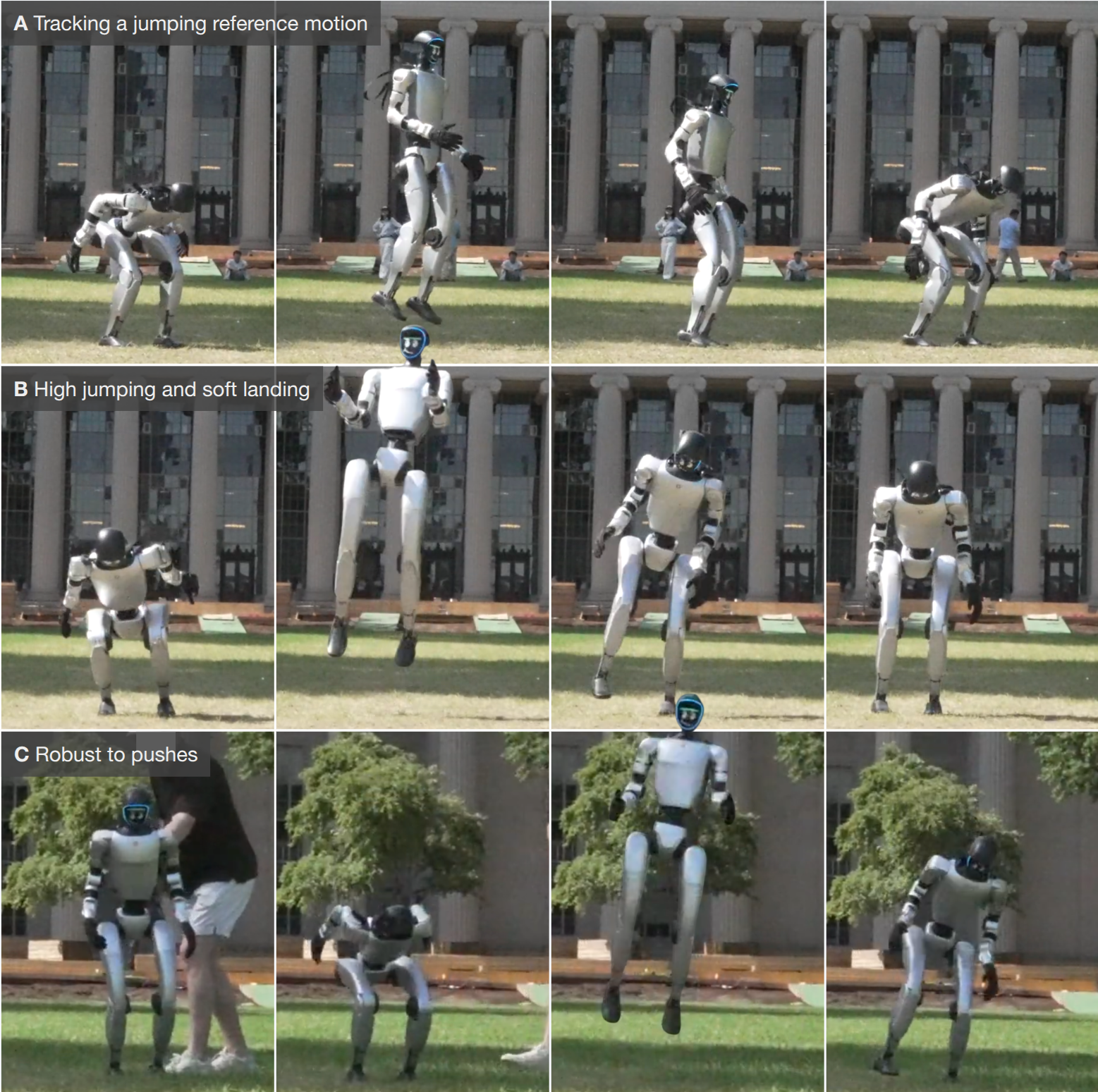}
    \caption{\textbf{Dynamic and robust jumping policies.} (A) The UAN policy first tracks a jumping reference motion, (B) then is fine-tuned to maximize jump height; (C) it is also robust to external disturbances.}
    \label{fig:jumping}
\end{figure}

\begin{figure}
        \centering
        \includegraphics[width=0.95\linewidth]{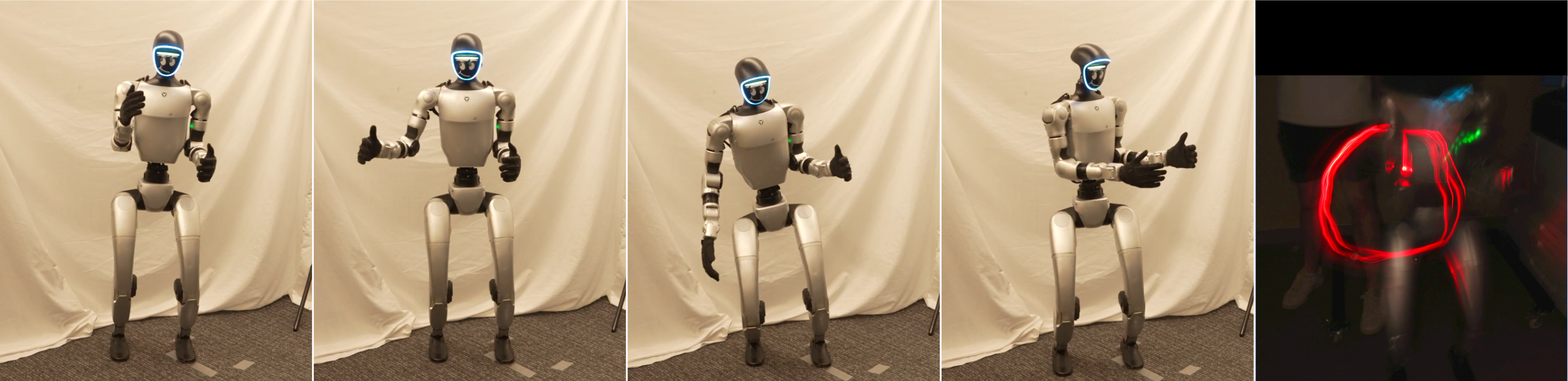}
        \caption{\textbf{Impedance position control.} Combining a precise force control policy with a proprioceptive state estimator, we develop an impedance controller for closed-loop end-effector position tracking. The controller tracks a circular trajectory reasonably well without any external sensing.}
        \label{fig:impedance}
\end{figure}    

\section{State Estimation}
\label{app:sec:state-estimation}

\subsection{Data Collection}
\label{app:ssec:se-data}
We collect data with the WTW-style data collection policy in its training
environment (Appendix~\ref{app:sec:wtw}), but on flat terrain with contact model parameters identical to those used in UAN training (Table \ref{tab:mujoco_contact_params}), for 20
simulation seconds with 4096 environments. The dataset is split 90\%/10\% into
training and validation samples by holding out a random 10\% of the
environments.

\subsection{Architectures}
\label{app:ssec:se-arch}
Table~\ref{tab:state-estimator-architectures} summarizes the estimator
architectures and shared optimization settings. Foot-height and
foot-projected-gravity estimators are trained separately for left-stance and
right-stance samples.

\begin{table}
\centering
\footnotesize
\caption{State-estimator architectures. All estimators use ELU activations, Adam, learning rate $3\times 10^{-4}$, batch size 4096, history length 3, and 200 epochs.}
\vspace{1em}
\label{tab:state-estimator-architectures}
\begin{tabular}{@{}lcccc@{}}
\toprule
Setting & Base velocity & Contact status & Foot height & Foot proj. gravity \\
\midrule
Output dimension & 3 & 2 & 1 per foot & 3 per foot \\
Hidden layers & $(256,256,256)$ & $(128,128)$ & $(256,256,256)$ & $(256,256,256)$ \\
Loss & MSE & BCE-logits & MSE & MSE \\
\bottomrule
\end{tabular}
\end{table}

\subsection{Observation Noise}
\label{app:ssec:se-noise}

The simulated observation-noise model used for estimator-data collection is
listed in Table~\ref{tab:state-estimator-noise}. The same noise model is used
when training the data-collection policy and the downstream policies except the
jumping task, which uses a smaller joint-velocity noise (Table~\ref{tab:jump-obs}).

\begin{table}[H]
\centering
\footnotesize
\caption{Observation noise used when collecting simulated data (uniform additive).}
\vspace{1em}
\label{tab:state-estimator-noise}
\begin{tabular}{@{}lc@{}}
\toprule
Observation & Noise range \\
\midrule
Base angular velocity & $\pm0.2~\mathrm{rad/s}$ \\
Projected gravity     & $\pm0.05$ \\
Joint position        & $\pm0.01~\mathrm{rad}$ \\
Joint velocity        & $\pm1.5~\mathrm{rad/s}$ \\
Previous target       & none \\
\bottomrule
\end{tabular}
\end{table}

\subsection{Contact-Weighted Root-Pose Fusion}
\label{app:state_estimation}

We initialize replay from a root pose whose simulated contacts reproduce the support wrench implied by the hardware trajectory. Since contact predictions are uncertain on hardware, we fuse pose estimates from all confident contacts rather than selecting a single stance foot.

Let $\hat{p}_{i,t}\in[0,1]$ be the estimated contact probability for foot $i\in\{L,R\}$, and let $p_{\mathrm{th}}$ be the contact-confidence threshold. We define contact weights
\begin{equation}
    w_{i,t}
    =
    \hat{p}_{i,t}\,\mathbf{1}[\hat{p}_{i,t}\geq p_{\mathrm{th}}],
    \qquad
    \alpha_{i,t}
    =
    \frac{w_{i,t}}{\sum_j w_{j,t} + \varepsilon}.
    \label{eq:contact_weights}
\end{equation}
States with no confident contact are excluded from the UAN initial-state buffer, since the stance-foot pose estimates cannot define a contact-consistent root pose.

Each confident contact provides an independent estimate of root height. Let $\hat{h}_{i,t}$ be the estimated height of foot $i$ relative to the terrain, and let $h^{\mathrm{fk}}_{i,t}(z=0)$ be the foot height from forward kinematics when the root height is set to zero. The root height implied by foot $i$ is
\begin{equation}
    \hat{z}^{(i)}_{t}
    =
    \hat{h}_{i,t} - h^{\mathrm{fk}}_{i,t}(z=0),
    \qquad
    \hat{z}_{t}
    =
    \sum\nolimits_i \alpha_{i,t}\,\hat{z}^{(i)}_{t}.
    \label{eq:contact_weighted_root_height}
\end{equation}

We recover root orientation using the same contact-weighted fusion. The orientation estimator predicts projected gravity in each stance-foot frame, $\hat{g}^{F_i}_{i,t}$. Forward kinematics maps each prediction to a root-frame candidate,
\begin{equation}
    \hat{g}^{(i)}_{t}
    =
    \operatorname{normalize}
    \!\left(
        R_{R \leftarrow F_i}(q_t)\,\hat{g}^{F_i}_{i,t}
    \right),
    \qquad
    \hat{g}_{t}
    =
    \operatorname{normalize}
    \!\left(
        \sum\nolimits_i \alpha_{i,t}\,\hat{g}^{(i)}_{t}
    \right).
    \label{eq:contact_weighted_root_orientation}
\end{equation}
On flat terrain, this projected-gravity estimate determines the roll and pitch used for replay initialization in the yaw-invariant local frame. Together with the fused root height $\hat{z}_t$ and the separately estimated base velocity, the resulting relabeled hardware state initializes replay with contact-consistent support forces.

\section{UAN Training Details}
\label{app:sec:uan-training}

Table~\ref{tab:g1-actuator-groups} lists the five G1 actuator groups over
which the UAN shares its per-group actor MLPs (Sec.~\ref{sec:uan_training}). Table \ref{tab:mujoco_contact_params} lists relevant contact model parameters used in our mjlab simulation~\cite{zakka2026mjlab}. 

\begin{table}[H]
\centering
\footnotesize
\caption{Unitree G1 actuator groups used by the shared UAN architecture.}
\vspace{1em}
\label{tab:g1-actuator-groups}
\begin{tabular}{@{}ll@{}}
\toprule
Actuator group & Joints \\
\midrule
\texttt{7520\_14} & Hip pitch, hip yaw, waist yaw \\
\texttt{7520\_22} & Hip roll, knee \\
\texttt{5020\_2x} & Waist roll, waist pitch, ankle pitch, ankle roll \\
\texttt{5020}     & Shoulder pitch, shoulder roll, shoulder yaw, elbow, wrist roll \\
\texttt{4010}     & Wrist pitch, wrist yaw \\
\bottomrule
\end{tabular}
\end{table}

\begin{table}[H]
\centering
\footnotesize
\caption{MuJoCo contact parameters used in the contact-pose sensitivity diagnostic.}
\vspace{1em}
\label{tab:mujoco_contact_params}
\begin{tabular}{@{}lcccc@{}}
\toprule
Element & condim & friction & solref & solimp \\
\midrule
Foot collision geoms
  & $3$
  & $[0.6,\,0.005,\,0.0001]$
  & $[0.02,\,1]$
  & $[0.9,\,0.95,\,0.001,\,0.5,\,2]$ \\
Terrain/ground geom
  & $3$
  & $[1,\,0.005,\,0.0001]$
  & $[0.02,\,1]$
  & $[0.9,\,0.95,\,0.001,\,0.5,\,2]$ \\
\bottomrule
\end{tabular}
\end{table}

Each replay episode corresponds to one labeled hardware window: 3 recorded
steps prime the UAN observation history, after which the simulator rolls the
window's 24 recorded joint-position targets forward with the UAN residual
applied. Windows are sampled only where the relabeled initial state has at
least one confident foot contact (contact probability $\geq 0.8$), and the
residual correction is clipped to $\pm 1.0$\,rad. The UAN replay reward terms
are listed in Table~\ref{tab:uan-reward}; PPO settings for the UAN, the
data-collection policy, and all downstream task policies are consolidated in
Table~\ref{tab:ppo-hparams}. We use the RSL-RL library~\cite{schwarke2025rslrl}.

\begin{table}
\centering
\footnotesize
\caption{PPO hyperparameters for the UAN, the WTW data-collection policy, and the downstream task policies.}
\vspace{1em}
\label{tab:ppo-hparams}
\resizebox{\linewidth}{!}{%
\begin{tabular}{@{}lccccc@{}}
\toprule
Parameter & UAN & WTW & Running & Force-control & Jumping \\
\midrule
\multicolumn{6}{@{}l}{\textit{Optimization}} \\
\textbf{Parallel environments} & \textbf{4096} & \textbf{4096} & \textbf{4096} & \textbf{4096} & \textbf{16384} \\
Steps per environment & 24 & 24 & 24 & 24 & 24 \\
Learning epochs       & 5 & 5 & 5 & 5 & 5 \\
Mini-batches          & 4 & 4 & 4 & 4 & 4 \\
Learning rate         & \multicolumn{5}{c}{$10^{-3}$, adaptive (target KL 0.01) for all} \\
Max gradient norm     & 1.0 & 1.0 & 1.0 & 1.0 & 1.0 \\
\addlinespace
\multicolumn{6}{@{}l}{\textit{PPO objective}} \\
Clip parameter         & 0.2 & 0.2 & 0.2 & 0.2 & 0.2 \\
\textbf{Entropy coefficient} & \textbf{0} & \textbf{0.002} & \textbf{0.01} & \textbf{0.002} & \textbf{0.005} \\
Value-loss coefficient & 1.0 & 1.0 & 1.0 & 1.0 & 1.0 \\
Use clipped value loss & yes & yes & yes & yes & yes \\
\textbf{Discount $\gamma$} & \textbf{0.95} & \textbf{0.99} & \textbf{0.99} & \textbf{0.99} & \textbf{0.99} \\
GAE $\lambda$          & 0.95 & 0.95 & 0.95 & 0.95 & 0.95 \\
\addlinespace
\multicolumn{6}{@{}l}{\textit{Networks (ELU activations)}} \\
Actor MLP  & per-group $(128,128)$ & $(512,512,256,128)$ & $(512,256,128)$ & $(512,512,256,128)$ & $(512,256,128)$ \\
Critic MLP & $(512,512,512,512)$ & $(512,512,512,512)$ & $(512,256,128)$ & $(512,512,512,512)$ & $(512,256,128)$ \\
\textbf{Initial action std.} & \textbf{0.3} & \textbf{1.0} & \textbf{1.0} & \textbf{1.0} & \textbf{1.0} \\
\bottomrule
\end{tabular}%
}
\end{table}

\begin{table}
\centering
\footnotesize
\caption{UAN replay reward weights.}
\vspace{1em}
\label{tab:uan-reward}
\begin{tabular}{@{}llr@{}}
\toprule
Term & Math & Weight \\
\midrule
\texttt{joint\_position\_mse} & $\frac{1}{n_q}\sum_j(q^{\mathrm{sim}}_{t,j}-q^{\mathrm{hw}}_{t,j})^2$ & $-1000$ \\
\texttt{joint\_velocity\_mse} & $\frac{1}{n_q}\sum_j(\dot{q}^{\mathrm{sim}}_{t,j}-\dot{q}^{\mathrm{hw}}_{t,j})^2$ & $-0.1$ \\
\texttt{residual\_rate\_l2} & $\frac{1}{n_q}\sum_j(\delta q_{t,j}-\delta q_{t-1,j})^2$ & $-0.01$ \\
\bottomrule
\end{tabular}
\end{table}

\section{Data Collection Policy Training Details}
\label{app:sec:wtw}

The data-collection policy is a WTW-style velocity-tracking policy
with behavior-shaping commands \cite{margolis2023walk}. In addition to the
base velocity command ($v^{\mathrm{cmd}}_x \in (-1.0, 1.0)$\,m/s,
$\omega^{\mathrm{cmd}}_z \in (-1.0, 1.0)$\,rad/s, no lateral command), the
policy receives a gait-frequency command in $(1.0, 2.5)$\,Hz and a
swing-height command in $(0.05, 0.1)$\,m; all commands are resampled every
5--10\,s. Varying these commands is what produces the diverse stance
durations, swing speeds, and impacts of the UAN training data
(Sec.~\ref{sec:uan_training}). In addition to standard velocity-tracking and
regularization terms, training adds gait-shaping penalties: a gait-phase
contact schedule ($-0.5$), a swing-height tracking term ($-20.0$), and a
Raibert-heuristic foot-placement term ($-5.0$, stance width $0.237$\,m). The
policy uses a 3-step observation history and is trained on procedurally
generated bumpy terrain; PPO settings are in Table~\ref{tab:ppo-hparams}.

To support initial hardware transfer before UAN training, we train the
data-collection policy in the nominal simulator with domain randomization.
At simulator startup, each environment samples a foot sliding-friction
coefficient uniformly from $[0.3,1.2]$, shared across its foot collision
geometries, independent joint target offsets uniformly within
$\pm0.015$\,rad, and torso center-of-mass offsets uniformly within
$\pm2.5$\,cm horizontally and $\pm3$\,cm vertically. These parameters
remain fixed for each environment throughout training. The policy is trained without a UAN.

\section{Downstream Policy Training Details}
\label{app:sec:downstream-training}

\textbf{Common setup.}
All downstream tasks share the same simulation and training setup. Simulation
runs at 200\,Hz (MuJoCo, $\Delta t = 5$\,ms) with a control decimation of 4,
so policies act at 50\,Hz; episodes last 20\,s. Actions are joint-position
offsets around the default standing configuration, applied to per-joint PD
controllers. During training, the frozen UAN deterministically adds its
residual correction to the commanded target before the simulator applies its
actuator and contact models; at deployment, only the downstream task policy is
exported. The observation-noise model is identical to that used during
state-estimation data collection (Table~\ref{tab:state-estimator-noise}),
except that the jumping task uses a smaller joint-velocity noise of
$\pm 0.5$\,rad/s (Table~\ref{tab:jump-obs}).

All tasks randomize foot--ground friction (uniform in $[0.3, 1.2]$, shared
across all foot geoms) and push the robot by perturbing its root velocity at
random intervals of 1--3\,s, with task-specific push ranges. For running and force control, policies trained
\emph{without} the UAN additionally randomize per-joint encoder bias (uniform
in $\pm 0.015$\,rad, applied to joint position observations) and the torso
center of mass (additive offset, uniform in $\pm 2.5$\,cm in $x$ and $y$,
$\pm 3$\,cm in $z$). Jumping uses the ranges in Appendix D.3. In the UAN-augmented simulator, all such robot-specific
randomization is removed (Sec.~\ref{sec:downstream_training}). PPO
hyperparameters for all tasks are listed in Table~\ref{tab:ppo-hparams}.

The running and jumping policies are trained with a torque-speed actuator
envelope~\cite{shin2023actuatorconstrained}. For joint $j$, denote stall torque by
$\bar{\tau}_j$ and no-load speed by $\bar{\dot{q}}_j$, and write
$r_j=\dot{q}_j/\bar{\dot{q}}_j$. The nominal PD torque satisfies
\begin{equation}
-\bar{\tau}_j\!\left[1+\max(\min(r_j,0),-1)\right]
\;\le\;\tau_j\;\le\;
\bar{\tau}_j\!\left[1-\max(\min(r_j,1),0)\right].
\label{eq:app-torque-speed-clipping}
\end{equation}
These asymmetric bounds retain full braking torque while reducing torque in the
direction of motion. Each control step, the commanded joint-position target is
clipped to satisfy these bounds, before adding the UAN residual in simulation;
deployment applies the same clipping.

For the task-policy reward tables, $I_{\mathrm{cmd}}$ denotes the
command-active mask, $I_i^c$ denotes contact for foot $i$, $I_i^{fc}$ denotes
first contact, $q^0$ denotes the default pose, and $\sigma_j(\cdot)$ denotes
the command-dependent posture tolerance.

\subsection{Running Policy}
\label{app:ssec:running}

To train a running policy, we adapt the default mjlab G1 velocity tracking
task~\cite{zakka2026mjlab} with several modifications. We remove the base
linear velocity observation to encourage the policy to rely on proprioception
rather than privileged state, and extend the observation history to 5 steps.
We add rough-terrain randomization and the torque-speed clipping of
Eq.~\eqref{eq:app-torque-speed-clipping} to improve robustness. To improve
motion quality, we lower the standing-command threshold for slow walking, add
a low-height termination at 0.5\,m to prevent kneeling local optima, and
increase the penalty weight on joint position limits to $-10$.
Table~\ref{tab:running-policy-reward} gives the reward terms.

On the command side, we remove heading-tracking environments and instead
dedicate 20\% of environments to pure turning. Commands follow a curriculum
over training iterations: $v^{\mathrm{cmd}}_x \in (-1.0, 1.0)$\,m/s and
$\omega^{\mathrm{cmd}}_z \in (-0.5, 0.5)$\,rad/s from iteration 0, widening to
$(-1.5, 2.0)$\,m/s and $(-0.7, 0.7)$\,rad/s at 5{,}000 iterations and to
$(-2.0, 3.0)$\,m/s at 10{,}000 iterations.

\begin{table}
\centering
\footnotesize
\caption{Running policy reward weights.}
\vspace{1em}
\label{tab:running-policy-reward}
\begin{tabular}{@{}llr@{}}
\toprule
Term & Math & Weight \\
\midrule
\multicolumn{3}{@{}l}{\textit{Task rewards}} \\
\texttt{track\_linear\_velocity}  & $\exp(-(\|v^{\mathrm{cmd}}_{xy}-v^b_{xy}\|^2 + (v^b_z)^2)/0.25)$ & $2.0$ \\
\texttt{track\_angular\_velocity} & $\exp(-(\omega^{\mathrm{cmd}}_z-\omega^b_z)^2/0.25)$ & $2.0$ \\
\addlinespace
\multicolumn{3}{@{}l}{\textit{Regularization}} \\
\texttt{upright} & $\exp(-\|g^b_{\mathrm{torso},xy}\|^2/0.2)$ & $1.0$ \\
\texttt{pose}    & $\exp\!\left(-\frac{1}{n_q}\sum_j\frac{(q_j-q^0_j)^2}{\sigma_j(\|v^{\mathrm{cmd}}\|)^2}\right)$ & $1.0$ \\
\addlinespace
\multicolumn{3}{@{}l}{\textit{Penalties}} \\
\texttt{dof\_pos\_limits}    & $\sum_j\left([q^{\min}_j-q_j]_+ + [q_j-q^{\max}_j]_+\right)$ & $-10.0$ \\
\texttt{foot\_clearance}     & $I_{\mathrm{cmd}}\sum_i |h_i-0.1|\,\|v^w_{i,xy}\|$ & $-2.0$ \\
\texttt{self\_collisions}    & $\sum_\ell \mathbf{1}\!\left[\max_k\|F^{\mathrm{self}}_{k,\ell}\|>10\right]$ & $-1.0$ \\
\texttt{foot\_swing\_height} & $I_{\mathrm{cmd}}\sum_i I_i^{fc}\left(h_i^{\mathrm{peak}}/0.1-1\right)^2$ & $-0.25$ \\
\texttt{foot\_slip}          & $I_{\mathrm{cmd}}\sum_i I_i^c\|v^w_{i,xy}\|^2$ & $-0.1$ \\
\texttt{action\_rate\_l2}    & $\sum_j(a_{t,j}-a_{t-1,j})^2$ & $-0.1$ \\
\texttt{angular\_momentum}   & $\|h\|^2$ & $-0.02$ \\
\texttt{soft\_landing}       & $I_{\mathrm{cmd}}\sum_i I_i^{fc}\|F_i\|$ & $-10^{-5}$ \\
\bottomrule
\end{tabular}
\end{table}

\subsection{Force Control Policy}
\label{app:ssec:force}

The commanded force is realized by a virtual force field: an impedance
spring--damper attached to the right wrist and anchored at the end-effector
position of the default standing pose. Every 10\,s, a target force is sampled
per axis from $\pm 30$\,N (zero force with probability 0.2) together with a
band stiffness $k \in (25, 400)$\,N/m and damping $0.1\,k$; horizontal force
commands are clipped so that the implied per-axis displacement $|F|/k$ stays
within a 0.3\,m reach. The virtual force is recomputed from the current end-effector
position and velocity and applied at every physics substep (200\,Hz), ramped
up over 2--4\,s, held for 2\,s, and ramped back down, with 2\,s of zero force
at the start of each episode; the applied force is clamped to 50\,N. The
policy observes the commanded force and uses a 20-step observation history.
The reward combines standing stabilization and force tracking
(Table~\ref{tab:force-control-policy-reward}), inspired by
\cite{portela2024learning}.

\begin{table}
\centering
\footnotesize
\caption{Force-control policy reward weights.}
\vspace{1em}
\label{tab:force-control-policy-reward}
\begin{tabular}{@{}llr@{}}
\toprule
Term & Math & Weight \\
\midrule
\multicolumn{3}{@{}l}{\textit{Task rewards}} \\
\texttt{track\_force\_ee}        & $\exp(-\|F^{\mathrm{cmd}}-F^{\mathrm{applied}}\|^2/50)$ & $3.0$ \\
\texttt{track\_lin\_vel\_xy\_exp} & $\exp(-\|v^b\|^2/0.25)$ & $2.0$ \\
\texttt{track\_ang\_vel\_z\_exp}  & $\exp(-\|\omega^b\|^2/0.5)$ & $2.0$ \\
\addlinespace
\multicolumn{3}{@{}l}{\textit{Regularization}} \\
\texttt{upright}     & $\exp(-\|g^b_{\mathrm{torso},xy}\|^2/0.2)$ & $1.0$ \\
\texttt{pose}        & $\exp\!\left(-\frac{1}{n_q}\sum_j\frac{(q_j-q^0_j)^2}{\sigma_j^2}\right)$ & $1.0$ \\
\addlinespace
\multicolumn{3}{@{}l}{\textit{Penalties}} \\
\texttt{dof\_pos\_limits}  & $\sum_j\left([q^{\min}_j-q_j]_+ + [q_j-q^{\max}_j]_+\right)$ & $-1.0$ \\
\texttt{self\_collisions}  & $\sum_\ell \mathbf{1}\!\left[\max_k\|F^{\mathrm{self}}_{k,\ell}\|>10\right]$ & $-1.0$ \\
\texttt{action\_rate\_l2}  & $\sum_j(a_{t,j}-a_{t-1,j})^2$ & $-0.1$ \\
\texttt{body\_ang\_vel}    & $\|\omega^w_{\mathrm{torso},xy}\|^2$ & $-0.05$ \\
\texttt{angular\_momentum} & $\|h\|^2$ & $-0.02$ \\
\bottomrule
\end{tabular}
\end{table}

\textbf{Impedance Control.}
As an additional experiment, we use the force-control policy for compliant
end-effector position control. We train a proprioceptive end-effector state
estimator with the setup of Sec.~\ref{app:sec:state-estimation}: it maps a history of onboard
observations (base angular velocity, projected gravity, joint positions and
velocities, and previous joint targets) to the right end-effector position
and velocity in the yaw-aligned base frame. At deployment, an outer impedance
loop converts the end-effector position error into a force command,
$F^{\mathrm{cmd}} = K_p(\hat{\mathbf{p}} - \mathbf{p}^{\mathrm{des}}) +
K_d\,\hat{\mathbf{v}}$ with $K_p = 50$\,N/m and $K_d = 0.2\,K_p$, which is fed
to the frozen force-control policy as its commanded force. The sign follows
the training convention: the policy realizes a commanded band force by
displacing the end-effector against it, so commanding a force pointing away
from $\mathbf{p}^{\mathrm{des}}$ drives the end-effector toward it. This realizes
compliant end-effector position control (e.g., tracking circular trajectories
of radius up to 0.3\,m) without any arm-specific controller.

\subsection{Jumping Policy}
\label{app:ssec:jumping}

\textbf{Overview.} 
The jumping task requires the G1 to perform repeated standing vertical jumps:
starting from rest, the robot must leave the ground with both feet, gain
height, and land softly back into a stable stance.  
We first train a reference-conditioned policy to track a retargeted jumping motion. 
The task is built on a
BeyondMimic-style motion tracking formulation~\cite{liao2025beyondmimic}.
We then fine-tune the policy with an additional reward for the predicted ballistic rise of the robot’s center of mass at takeoff. The motion-tracking rewards retain their original weights during fine-tuning, while reference-error termination conditions are replaced with general fall conditions, allowing the policy to increase jump height while preserving the reference motion’s structure.
The reference motion is the jump from CMU MoCap clip 02\_04 (``jump and
balance'')~\cite{cmumocap}, retargeted to the G1 and resampled at 50\,Hz (200
frames, 4\,s). The clip is padded with 1\,s of standing at the beginning and
end (300 frames, 6\,s total), so the policy must initiate the jump from rest
and hold a stable stance after landing.

\textbf{Episode structure and reference-state initialization.}
The initial reference frame is sampled using failure statistics accumulated over temporal bins, followed by uniform sampling within the selected bin.
The corresponding reference state is
perturbed with uniform noise on the root pose ($\pm 5$\,cm in $xy$,
$\pm 1$\,cm in $z$, $\pm 0.1$\,rad roll/pitch, $\pm 0.2$\,rad yaw), root
velocity ($\pm 0.5$\,m/s linear $xy$, $\pm 0.2$\,m/s vertical,
$\pm 0.52$\,rad/s roll/pitch, $\pm 0.78$\,rad/s yaw), and joint positions
($\pm 0.1$\,rad). 

\textbf{Observations and actions.}
The actor combines onboard proprioception with features of the current reference frame; it does not observe the robot’s global position or measured root linear velocity. Table~\ref{tab:jump-obs} lists the observation
terms; concatenated, they form a 161-dimensional policy input. The reference
joint targets in the \emph{command} term provide the policy with an implicit
phase signal for the clip. The critic additionally observes privileged,
noise-free state: the root linear velocity, the reference anchor position and orientation expressed in the base frame, and the positions and orientations of all 14
tracked bodies relative to the robot's anchor frame. Actions
follow the common setup with a per-joint action scale of
$0.25\,\tau^{\max}_j / k_{p,j}$, where $\tau^{\max}_j$ and $k_{p,j}$ are the
effort limit and proportional gain of joint $j$~\cite{liao2025beyondmimic}, and the torque--speed
envelope of Eq.~\eqref{eq:app-torque-speed-clipping}.

\begin{table}
  \centering
  \footnotesize
  \caption{Actor observation space for the jumping task. Noise is
  additive, sampled uniformly from $[-n, n]$ per component. The anchor
  body is the torso link.}
  \vspace{1em}
  \label{tab:jump-obs}
  \begin{tabular}{@{}lcc@{}}
    \toprule
    Term & Dim. & Noise $n$ \\
    \midrule
    Reference joint positions and velocities (current frame) & 58 & -- \\
    Reference height & 1 & -- \\
    Reference projected gravity & 3 & -- \\
    Reference root velocity & 6 & -- \\
    Projected gravity (IMU) & 3 & 0.05 \\
    Base angular velocity (IMU) & 3 & 0.2 \\
    Joint positions (relative to default) & 29 & 0.01 \\
    Joint velocities & 29 & 0.5 \\
    Previous action & 29 & -- \\
    \midrule
    Total & 161 & \\
    \bottomrule
  \end{tabular}
\end{table}

\textbf{Rewards.}
Table~\ref{tab:jump-rewards} lists all reward terms. Let $\hat{(\cdot)}$
denote reference quantities, $\mathbf{p}_a$ and $\mathbf{q}_a$ the position
and orientation of the anchor body (torso), $\mathcal{B}$ the set of 14
tracked bodies, and $\mathbf{q}_1 \ominus \mathbf{q}_2$ the geodesic angle
between two orientations. Before computing body position and orientation
errors, the reference bodies are rigidly re-anchored to the robot's current
anchor $xy$-position and yaw heading, making these terms invariant to global
horizontal drift. The jump bonus uses the whole-body center-of-mass vertical velocity at takeoff to estimate ballistic rise, \(h_{\mathrm{ballistic}}=\max(v_{\mathrm{COM},z}^{\mathrm{takeoff}},0)^2/(2g)\). A single bonus is available per reference-motion segment, provided the segment begins with a reference frame classified as non-airborne. The robot must establish two-foot ground contact for two consecutive policy steps, followed by two consecutive steps with both feet out of contact. The velocity is recorded at the first airborne step, and additional hops within the same segment receive no further bonus.

\begin{table}
\centering
\footnotesize
\caption{Reward terms for jumping pretraining and fine-tuning.
Motion-tracking and regularization weights remain unchanged during
fine-tuning, which adds a ballistic center-of-mass rise bonus. The indicator $\mathbf{1}_{\mathrm{to}}$ marks a confirmed
two-foot takeoff, rewarded at most once per eligible reference-motion
segment. Here $[x]_+ = \max(x,0)$, and $q_j^{\min}$ and $q_j^{\max}$
are the soft joint-position limits. All terms are multiplied by the
policy timestep $\Delta t_{\rm policy}$ when computing the reward.}
\vspace{1em}
\label{tab:jump-rewards}
\begin{tabular}{@{}llcc@{}}
\toprule
Term & Math & Pretraining & Fine-tuning \\
\midrule
\multicolumn{4}{@{}l}{\textit{Motion tracking}} \\
\texttt{motion\_global\_root\_pos}
& $\exp\!\big(-\|\hat{\mathbf{p}}_a-\mathbf{p}_a\|^2/0.3^2\big)$
& $0.5$ & $0.5$ \\
\texttt{motion\_global\_root\_ori}
& $\exp\!\big(-(\hat{\mathbf{q}}_a\ominus\mathbf{q}_a)^2/0.4^2\big)$
& $0.5$ & $0.5$ \\
\texttt{motion\_body\_pos}
& $\exp\!\big(-\tfrac{1}{|\mathcal{B}|}\sum_b
  \|\hat{\mathbf{p}}_b-\mathbf{p}_b\|^2/0.3^2\big)$
& $1.0$ & $1.0$ \\
\texttt{motion\_body\_ori}
& $\exp\!\big(-\tfrac{1}{|\mathcal{B}|}\sum_b
  (\hat{\mathbf{q}}_b\ominus\mathbf{q}_b)^2/0.4^2\big)$
& $1.0$ & $1.0$ \\
\texttt{motion\_body\_lin\_vel}
& $\exp\!\big(-\tfrac{1}{|\mathcal{B}|}\sum_b
  \|\hat{\mathbf{v}}_b-\mathbf{v}_b\|^2/1.0^2\big)$
& $1.0$ & $1.0$ \\
\texttt{motion\_body\_ang\_vel}
& $\exp\!\big(-\tfrac{1}{|\mathcal{B}|}\sum_b
  \|\hat{\boldsymbol{\omega}}_b-\boldsymbol{\omega}_b\|^2/3.14^2\big)$
& $1.0$ & $1.0$ \\
\addlinespace
\multicolumn{4}{@{}l}{\textit{Jump objective}} \\
\texttt{jump\_com\_ballistic\_rise}
& $\mathbf{1}_{\mathrm{to}}\,
  \dfrac{[v_{\mathrm{COM},z}^{\mathrm{takeoff}}]_+^2}
  {2g\,\Delta t_{\rm policy}}$
& $0$ & 5.0 \\
\addlinespace
\multicolumn{4}{@{}l}{\textit{Regularization}} \\
\texttt{action\_rate\_l2}
& $\|\mathbf{a}_t-\mathbf{a}_{t-1}\|^2$
& $-0.1$ & $-0.1$ \\
\texttt{joint\_limit}
& $\sum_j \big([q_j^{\min}-q_j]_+ + [q_j-q_j^{\max}]_+\big)$
& $-10$ & $-10$ \\
\texttt{self\_collisions}
& \# substeps with self-contact force $>10$\,N
& $-10$ & $-10$ \\
\bottomrule
\end{tabular}
\end{table}

\textbf{Terminations.}
During pretraining, reference-error terminations encourage accurate motion tracking. During fine-tuning, these conditions are replaced by minimum-root-height and maximum-root-tilt conditions, allowing increased jump height without terminating solely because the policy departs from the reference. Both stages use a 20-s episode timeout.

\begin{table}
\centering
\footnotesize
\caption{Termination conditions for jumping pretraining and fine-tuning.
Thresholds are fixed within each stage; a dash indicates an inactive
condition. The set $\mathcal{E}$ contains the tracked feet and wrists,
$g_{a,z}$ is the vertical component of projected gravity in the anchor
frame, and $\alpha_{\mathrm{root}}$ is the root tilt angle from upright.}
\vspace{1em}
\label{tab:jump-terminations}
\begin{tabular}{@{}llcc@{}}
\toprule
Condition & Trigger & Pretraining & Fine-tuning \\
\midrule
Episode timeout
& $t_{\mathrm{episode}} \ge \theta$
& $20$\,s & $20$\,s \\
Anchor height error
& $|\hat{z}_a-z_a|>\theta$
& $0.25$\,m & -- \\
Anchor gravity mismatch
& $|\hat{g}_{a,z}-g_{a,z}|>\theta$
& $0.8$ & -- \\
End-effector height error
& $\max_{b\in\mathcal{E}}|\hat{z}_b-z_b|>\theta$
& $0.25$\,m & -- \\
Minimum root height
& $z_{\mathrm{root}}<\theta$
& -- & $0.30$\,m \\
Maximum root tilt
& $\alpha_{\mathrm{root}}>\theta$
& -- & $80^\circ$ \\
\bottomrule
\end{tabular}
\end{table}

\textbf{Domain randomization.}
Jumping uses the foot--ground friction randomization and disturbance
intervals described in the common setup. Additive root-velocity
disturbances are applied throughout the motion, with uniform ranges of
$\pm0.5$\,m/s horizontally, $\pm0.2$\,m/s vertically,
$\pm0.52$\,rad/s in roll and pitch, and $\pm0.78$\,rad/s in yaw.
For jumping, the baseline-specific randomization uses joint encoder
biases within $\pm0.01$\,rad and torso center-of-mass offsets within
$\pm2.5$\,cm along $x$ and $\pm5$\,cm along $y$ and $z$.
These parameters are sampled uniformly at simulator startup and remain
fixed for each environment throughout training.

\section{Sim-to-Sim Experiment Details}
\label{app:sec:sim2sim_setup}

\textbf{Target simulator construction.}
For sim-to-sim evaluation, we construct \texttt{easy}, \texttt{medium}, and \texttt{hard} target simulators by applying deterministic actuator perturbations to the nominal G1 model. The perturbations modify joint damping, frictionloss, action delay, encoder bias, and armature. The \texttt{easy} target is chosen to produce a measurable but recoverable tracking gap for the nominal baseline; \texttt{medium} and \texttt{hard} scale the same perturbation vector to test increasingly severe actuator mismatch. The values are listed in Table~\ref{tab:sim2sim_targets}.

\begin{table}[h]
\centering
\caption{\textbf{Sim-to-sim target perturbations.} Each target simulator applies deterministic actuator perturbations to the nominal G1 model. The \texttt{medium} and \texttt{hard} targets scale the same perturbation vector used for \texttt{easy}.}
\label{tab:sim2sim_targets}
\renewcommand{\arraystretch}{1.1}
\begin{tabular}{lccc}
\toprule
\textbf{Perturbation} & \textbf{\texttt{easy}} & \textbf{\texttt{medium}} & \textbf{\texttt{hard}} \\
\midrule
Damping offset & $+2.0$ & $+4.0$ & $+6.0$ \\
Frictionloss offset & $+0.5$ & $+1.0$ & $+1.5$ \\
Action delay & $1$ step & $2$ steps & $3$ steps \\
Encoder bias & $+0.003$ rad & $+0.006$ rad & $+0.009$ rad \\
Armature multiplier & $1.5\times$ & $2.0\times$ & $2.5\times$ \\
\bottomrule
\end{tabular}
\end{table}

\textbf{Sim-to-sim policy training.}
For each target simulator, we train downstream velocity-tracking policies either in the nominal simulator or in the UAN-augmented simulator. The WTW UAN is trained from $5$ minutes of WTW target data. The WTW+Vel UAN is trained from the same $5$ minutes of WTW data plus $5$ minutes of running data, allowing us to test whether task-specific running data improves over the walking-only residual model.

\section{World Model Details}
\label{app:sec:rwm}

\textbf{Overview and scope.} We compare Contact-UAN with our best-effort G1
adaptation of the fully learned offline RWM-U~\cite{li2025uncertainty} on a
single \texttt{hard} target environment for omnidirectional base velocity
tracking on flat ground. This is a controlled data-efficiency reference,
not a definitive benchmark of the original method.

\textbf{Data-efficiency results.} A policy trained in the nominal physics
simulator collects data in the \texttt{hard} target environment for model
learning. In Fig.~\ref{fig:sim2sim_data_efficiency}, UAN trained with 5 minutes
of target data reaches the expert policy's normalized episode length
(100\%). Our fully learned world model uses approximately 33 hours of
target data (about $400\times$ as much): a policy initialized from scratch
inside it reaches 6.8\% normalized episode length on the target, and a
policy fine-tuned from the data-collection policy reaches 38.2\%, compared
with 32.6\% for the nominal-simulator baseline. All policies reach 100\%
in their training environments. These results support the value of the
simulator's physics priors in this controlled setting.

\input{figures/camera_ready/world_model_figure}

\textbf{Observation, action, reward.} The policy receives the same
$99$-dimensional observation as the on-policy baseline: base linear and
angular velocity read from the IMU site \texttt{imu\_in\_pelvis} ($6$),
projected gravity ($3$), a $3$-D commanded base velocity, joint position
residuals $q-q_{\text{default}}$ ($29$), joint velocities ($29$), and the
previous action ($29$). 

\textbf{Data collection.} We collect $6$ million transitions (around 33 hours of data) on the \texttt{hard} target environment using a baseline policy. For data collection we record the IMU sensor readings rather than the body-state \texttt{lin\_vel\_w}/\texttt{ang\_vel\_w} pair, so
the world model is trained on exactly the inputs the policy sees at
evaluation time.

\textbf{Robotic world model.} The world model is a GRU ensemble ($5$
members, hidden size $256$, $2$ stacked layers) with a $32$-step history
horizon, an $8$-step forecast horizon, and a contact channel of dimension
$30$ (Table~\ref{tab:rwmu-wm}). At policy-training time, a per-step
epistemic-uncertainty signal formed from ensemble disagreement on next-state
predictions is subtracted from the reward with MOPO penalty $\lambda = 1.0$ \cite{li2025uncertainty}.
Imagined rollouts terminate when the learned termination head exceeds $0.9$;
lower thresholds collapsed training because the policy
learned to exploit short, high-reward fragments inside the model.

\textbf{Policy training.} Given a trained world model, we run PPO on
imagined rollouts using the same \texttt{rsl\_rl} algorithm and
hyperparameters as the baseline (Table~\ref{tab:rwmu-ppo}). The outer loop
runs for $5{,}000$ iterations, following \cite{li2025uncertainty}; training
longer overfits the policy to the inaccurate WM and degrades target
performance. UAN is trained using only 5 minutes of data.

\textbf{Paradigms compared.} The \emph{offline} world model is trained
once on the $6$\,M-transition dataset collected by the baseline policy.
Variant WM trains a randomly initialized actor inside this learned model.
Variant WM FT reuses the same learned world model but fine-tunes the actor
from the data-collection policy checkpoint and adds a behavior anchor
$\beta\,\mathrm{KL}(\pi\Vert\pi_{\text{baseline}})$ with $\beta\!=\!1.0$ to
the PPO loss. Thus, ``FT'' refers to policy fine-tuning within the world
model, not to fine-tuning the world model itself.

\textbf{Evaluation protocol.} Every policy is evaluated on the \texttt{hard} target with $32$ parallel
environments, $2{,}000$ simulator steps (up to two $1{,}000$-step episodes per
environment), and a $20$\,s episode cap. We report mean and standard deviation
of \emph{survival length} (\%), defined as episode length divided by the
$1{,}000$-step cap, across all completed episodes. The
evaluation environment and seed distribution are identical
across Expert, Baseline, UAN, and all RWM-U variants.
This normalized mean episode length is defined separately from the 20-s survival
percentages reported for the reset and sensor-corruption experiments in
Appendix~\ref{app:additional_ablation}.

\begin{table}
\centering
\footnotesize
\caption{World-model architecture and training hyperparameters used for RWM-U on G1.}
\vspace{1em}
\label{tab:rwmu-wm}
\begin{tabular}{@{}lc@{}}
\toprule
Parameter & Value \\
\midrule
\multicolumn{2}{@{}l}{\textit{Architecture}} \\
Backbone              & GRU ensemble \\
Ensemble size         & 5 \\
RNN hidden size       & 256 \\
RNN layers            & 2 \\
Head MLP (state / ext. / contact / term.) & $[128]$ \\
\addlinespace
\multicolumn{2}{@{}l}{\textit{Dimensions}} \\
History horizon $H$   & 32 \\
Forecast horizon $F$  & 8 \\
Observation dim       & 99 \\
Action dim            & 29 \\
Contact channel dim   & 30 \\
\addlinespace
\multicolumn{2}{@{}l}{\textit{Data}} \\
Dataset size          & 6M transitions $\approx$ 33 hours \\
File shard size       & 10{,}000 rows \\
\addlinespace
\multicolumn{2}{@{}l}{\textit{Optimization}} \\
Batch size            & 1024 \\
Epochs                & 4 \\
Optimizer             & Adam \\
Learning rate         & $1\!\times\!10^{-4}\!\to\!1\!\times\!10^{-6}$ (cosine) \\
Weight decay          & $1\!\times\!10^{-5}$ \\
Gradient steps (approx.) & $1.8\!\times\!10^{4}$ \\
\bottomrule
\end{tabular}
\end{table}

\begin{table}
\centering
\footnotesize
\caption{PPO hyperparameters used for MOPO-PPO policy training inside the RWM-U world model. The baseline PPO column is shown for reference; only the bold rows differ.}
\vspace{1em}
\label{tab:rwmu-ppo}
\begin{tabular}{@{}lcc@{}}
\toprule
Parameter & Baseline PPO & MOPO-PPO (RWM-U) \\
\midrule
\multicolumn{3}{@{}l}{\textit{Rollout collection}} \\
Parallel envs          & 4096 & 4096 \\
Steps per environment  & 24 & 24 \\
\textbf{Rollout source} & \textbf{simulator} & \textbf{learned WM} \\
\textbf{Outer iterations} & \textbf{20{,}000} & \textbf{5{,}000} \\
\addlinespace
\multicolumn{3}{@{}l}{\textit{Optimization}} \\
Learning epochs        & 5 & 5 \\
Mini-batches           & 4 & 4 \\
Learning rate          & $10^{-3}$ & $10^{-3}$ \\
Schedule               & adaptive & adaptive \\
Max gradient norm      & 1.0 & 1.0 \\
\addlinespace
\multicolumn{3}{@{}l}{\textit{PPO objective}} \\
Clip parameter         & 0.2 & 0.2 \\
\textbf{Entropy coefficient} & \textbf{0.01} & \textbf{0.008} \\
Value-loss coefficient & 1.0 & 1.0 \\
Use clipped value loss & yes & yes \\
Discount $\gamma$      & 0.99 & 0.99 \\
GAE $\lambda$          & 0.95 & 0.95 \\
Target KL              & 0.01 & 0.01 \\
Actor / critic MLP     & $[256,128,128]$ & $[256,128,128]$ \\
\addlinespace
\multicolumn{3}{@{}l}{\textit{RWM-U specific}} \\
\textbf{Uncertainty penalty $\lambda$} & {--} & \textbf{1.0} \\
\textbf{Termination threshold}         & {--} & \textbf{0.9} \\
\bottomrule
\end{tabular}
\end{table}

\section{Additional Ablations and Robustness Experiments}
\label{app:additional_ablation}

\subsection{Reset and Estimator Protocols}
\label{app:reset_protocol}

\textbf{Training under different reset conditions.}
The reset-training ablation uses the \texttt{hard} target and compares
UANs and downstream policies trained with direct, estimated,
component-oracle, or full-oracle resets. The component oracles replace
estimated base velocity (Base), contact state (Cnt.), or stance-foot pose
(Pose) with simulator ground truth; the full oracle (All) replaces all
three. The pose being initialized is the simulator-side contact pose,
including the stance penetration and orientation needed for contact-consistent
replay. The reported 20-s survival rises from 49.8\% with direct reset
to 96.7\% with estimated reset; component/full-oracle variants yield
96.6--98.4\%. Linear and yaw tracking errors differ by less than
approximately 10\% between estimated and oracle variants.
The separately trained direct, estimated, and full-oracle conditions have
24-step joint-position replay RMSEs of 0.0318, 0.0239, and 0.0120\,rad,
respectively.

\textbf{Sensitivity analysis.}
A separate experiment evaluates the \emph{same} residual with estimated
and ground-truth initialization, either enabled or disabled. With estimated
initialization, enabling the residual reduces 24-step joint-position RMSE
from 0.0750 to 0.0239\,rad; with ground-truth initialization it reduces
RMSE from 0.0720 to 0.0184\,rad (Figure~\ref{fig:state_estimation}C). Figure~\ref{fig:reset_sensitivity_full}
shows the corresponding joint-velocity errors. Even with ground-truth initialization, enabling the UAN reduces error. Because this improvement persists when initialization error is removed, it supports the interpretation that the UAN learns actuator-dynamics corrections rather than merely compensating for errors in the estimated initial state.

\begin{figure}[ht]
    \centering
    \begin{tikzpicture} \begin{axis}[scale only axis,width=99pt,height=2.6cm,xmin=-0.16,xmax=1.16,ymin=0,ymax=1.08,xtick={0,1},xticklabels={{\shortstack{Without\\residual}},{\shortstack{With\\residual}}},xticklabel style={align=center,yshift=-4pt},ytick={0,0.5,1},yticklabels={0,0.5,1.0},ylabel={$\dot{q}$ RMSE (rad/s)},tick label style={font=\scriptsize},label style={font=\scriptsize},tick align=inside,xtick pos=bottom,scaled y ticks=false,legend style={font=\scriptsize,draw=none,at={(0.5,1.02)},anchor=south,legend columns=2,/tikz/every even column/.append style={column sep=3pt}},clip=false] \addplot[BaselineGray!85!black,densely dashed,mark=o,mark size=2pt,line width=0.7pt] coordinates {(0,0.9663720533) (1,0.4000482456)}; \addlegendentry{Ground truth} \addplot[UANTeal,mark=square*,mark size=2pt,line width=0.7pt] coordinates {(0,0.9937351860) (1,0.4446492440)}; \addlegendentry{Estimated} \end{axis} \end{tikzpicture}
    \caption{\textbf{Fixed-residual initialization sensitivity: joint velocity.} 24-step joint-velocity replay RMSE in rad/s in the sim-to-sim experiment. The same residual is either disabled or enabled under ground-truth and estimated initialization. Joint-position results are shown in Figure~\ref{fig:state_estimation}C.}
    \label{fig:reset_sensitivity_full}
\end{figure}

\subsection{Retraining with Structured Sensor Corruption}
\label{app:sensor_corruption}

We retrain with data corrupted by fixed IMU pitch biases of
$0.5^\circ$, $1^\circ$, and $2^\circ$, or causal velocity low-pass
filtering at 20, 10, and 5\,Hz. The filtering includes its associated
delay. Across these conditions, reported 20-s survival remains
97.1--100\% (Fig.~\ref{fig:sensor_corruption}), 24-step joint-position and
joint-velocity replay RMSEs remain within 8\% and 11\% of clean,
respectively, and command-tracking errors remain within approximately
15\% of clean. 

\begin{figure}[ht]
    \centering
    \begin{tikzpicture} \definecolor{CorruptionOrange}{HTML}{D55E00} \begin{axis}[scale only axis,width=240pt,height=2.6cm,ybar,bar width=15pt,bar shift=0pt,xmin=0.35,xmax=8.45,ymin=0,ymax=110,xtick={1,2.4,3.4,4.4,5.8,6.8,7.8},xticklabels={Clean,0.5,1,2,20,10,5},ytick={0,50,100},ylabel={20-s survival (\%)},tick label style={font=\scriptsize},label style={font=\scriptsize},xticklabel style={align=center,yshift=-4pt},tick align=inside,xtick pos=bottom,xtick style={draw=none},scaled y ticks=false,clip=false] \addplot+[fill=BaselineGray,draw=BaselineGray,bar shift=0pt,error bars/.cd,y dir=both,y explicit,error mark=-,error bar style={draw=BaselineGray!80!black,line width=0.35pt},error mark options={draw=BaselineGray!80!black,line width=0.35pt,rotate=90,mark size=2pt}] table[x=x,y=value,y error minus=err_low,y error plus=err_high,row sep=\\] {x value err_low err_high\\ 1 100.00000000 0.09369599 0.00000000\\ }; \addplot+[fill=UANTeal,draw=UANTeal,bar shift=0pt,error bars/.cd,y dir=both,y explicit,error mark=-,error bar style={draw=BaselineGray!80!black,line width=0.35pt},error mark options={draw=BaselineGray!80!black,line width=0.35pt,rotate=90,mark size=2pt}] table[x=x,y=value,y error minus=err_low,y error plus=err_high,row sep=\\] {x value err_low err_high\\ 2.4 97.11914258 0.55805914 0.46975967\\ 3.4 100.00000000 0.09369599 0.00000000\\ 4.4 99.75585987 0.20471142 0.11147231\\ }; \addplot+[fill=CorruptionOrange,draw=CorruptionOrange,bar shift=0pt,error bars/.cd,y dir=both,y explicit,error mark=-,error bar style={draw=BaselineGray!80!black,line width=0.35pt},error mark options={draw=BaselineGray!80!black,line width=0.35pt,rotate=90,mark size=2pt}] table[x=x,y=value,y error minus=err_low,y error plus=err_high,row sep=\\] {x value err_low err_high\\ 5.8 100.00000000 0.09369599 0.00000000\\ 6.8 99.97558672 0.11375668 0.02010491\\ 7.8 100.00000000 0.09369599 0.00000000\\ }; \node[anchor=south,font=\scriptsize] at (rel axis cs:0.37654321,1.04) {IMU bias ($^\circ$)}; \node[anchor=south,font=\scriptsize] at (rel axis cs:0.79629630,1.04) {Velocity LPF (Hz)}; \end{axis} \end{tikzpicture}
    \caption{\textbf{Sensor-corruption retraining experiment.} Reported
    20-s survival after training with clean data, fixed IMU pitch bias
    (degrees), or velocity low-pass filtering (Hz).}
    \label{fig:sensor_corruption}
\end{figure}

\subsection{Matched Actuator-Model Comparison}
\label{app:actuator_baselines}

Table~\ref{tab:actuator_baselines} compares a position residual, a torque
residual, and a torque-supervised actuator model in a matched 200-Hz
simulation experiment. Torque supervision is available in this simulation comparison;
it does not imply reliable output-torque sensing on G1 hardware.
The three models show similar downstream performance.

\begin{table}[ht]
\centering
\caption{\textbf{Matched actuator-model comparison.} While the position residual UAN achieves the tightest fit to target data, all variations achieve similar downstream performance.}
\label{tab:actuator_baselines}
\begin{tabular}{@{}lccccc@{}}
\toprule
Model & Replay $q$ & Survival & XY error & Yaw error & Power \\
 & (rad) & (\%) & (m/s) & (rad/s) & (W) \\
\midrule
Position residual & 0.00349 & 97.34 & 0.391 & 0.185 & 69.29 \\
Torque residual & 0.00571 & 98.17 & 0.387 & 0.214 & 69.25 \\
Supervised torque & 0.00565 & 97.56 & 0.394 & 0.215 & 71.90 \\
\bottomrule
\end{tabular}
\end{table}

\subsection{Additional Replay Visualizations}
\label{app:replay_visualizations}

\begin{figure}[ht]
    \centering
    \includegraphics[width=0.95\linewidth]{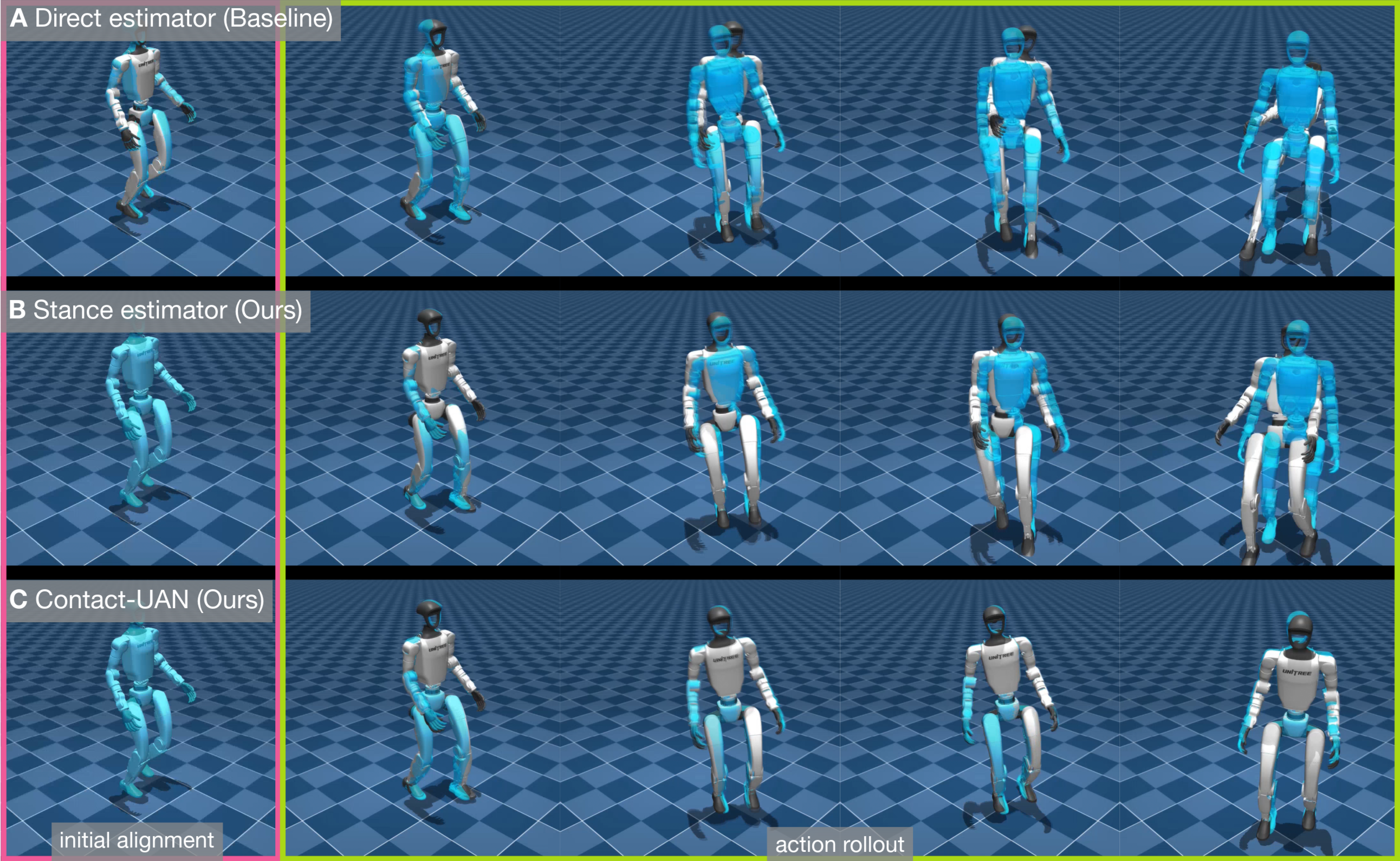}
    \caption{\textbf{Open-loop rollouts.} We initialize the simulator with
    the estimated state and roll out recorded hardware actions. The
    estimated simulator reset state is shown in blue. (A) uses a foot-height
    estimator trained on both stance and swing data; (B) uses our estimator
    trained only on stance data to align initial contacts; (C) uses our
    estimator and UAN to reduce the actuation gap.}
    \label{fig:open_loop_rollout}
\end{figure}

Figure~\ref{fig:knee_position_replay} shows representative knee-position trajectories on held-out hardware running data. Figure~\ref{fig:open_loop_rollout} shows the qualitative open-loop replay sequence. The horizontal knee-torque/aggregate-usage comparison is in Figure~\ref{fig:running_torque_gap}.

\input{figures/camera_ready/appendix_knee_position}

%% file: figures/camera_ready/world_model_figure.tex
\definecolor{ExpertOrange}{HTML}{D55E00}
\begin{figure}[ht]
    \centering
    \begin{minipage}{0.74\linewidth}

    \centering
    \begin{tikzpicture}
        \node[inner sep=0pt] at (0,0) {
            \scriptsize
            \begin{tabular}{@{}cc@{}}
                \tikz{\draw[fill=UANTeal, draw=UANTeal] (0,0) rectangle (0.12,0.12);} Train Env &
                \tikz{\draw[fill=ExpertOrange, draw=ExpertOrange] (0,0) rectangle (0.12,0.12);} Target Env
            \end{tabular}
        };
    \end{tikzpicture}
    
    \vspace{0.2em}

 \begin{tikzpicture}
\begin{axis}[
    width=\linewidth,
    height=4.8cm,
    ybar,
    bar width=6pt,
    ymin=0,
    ymax=115,
    ytick={0,25,50,75,100},
    enlarge x limits=0.18,
    xtick={1,2,3,4,5},
xticklabels={
    {\shortstack{Expert}},
    {\shortstack{Base}},
    {\shortstack{WM\\33 h}},
    {\shortstack{WM FT\\33 h}},
    {\shortstack{UAN\\5 min}}
},
    x tick label style={font=\scriptsize, align=center},
    xtick pos=bottom,
    ylabel={Normalized episode length (\%)},
    tick label style={font=\scriptsize},
    label style={font=\scriptsize},
]
        \addplot+[
            fill=UANTeal,
            draw=UANTeal,
            bar shift=-3.7pt,
            error bars/.cd,
            y dir=both,
            y explicit,
            error mark=-,
            error bar style={draw=UANTeal!80!black, line width=0.35pt},
            error mark options={draw=UANTeal!80!black, line width=0.35pt, rotate=90, mark size=2.0pt},
        ] coordinates {
            (1,100) +- (0,0.0)
            (2,100) +- (0,0.0)
            (3,100) +- (0,0.0)
            (4,100) +- (0,0.0)
            (5,100) +- (0,0.0)
        };

        \addplot+[
            fill=ExpertOrange,
            draw=ExpertOrange,
            bar shift=3.7pt,
            error bars/.cd,
            y dir=both,
            y explicit,
            error mark=-,
            error bar style={draw=ExpertOrange!80!black, line width=0.35pt},
            error mark options={draw=ExpertOrange!80!black, line width=0.35pt, rotate=90, mark size=2.0pt},
        ] coordinates {
            (1,100.0) +- (0,0.0)
            (2,32.6) +- (0,25.8)
            (3,6.8)   +- (0,1.0)
            (4,38.2)  +- (0,31.6)
            (5,100.0) +- (0,0.0)
        };

        \node[
            anchor=north,
            font=\tiny,
            inner sep=1pt,
            fill=white,
            fill opacity=0.75,
            text opacity=1
        ] at (rel axis cs:0.5,0.98) {$\uparrow$ better};
    \end{axis}
\end{tikzpicture}

    \vspace{0.2em}

    \end{minipage}
    \caption{\textbf{Controlled world-model data-efficiency reference.}
    Mean normalized episode length with standard deviation in each method's
    training environment and the \texttt{hard} target environment.
    UAN uses 5 minutes of target data; our best-effort G1 adaptation of
    RWM-U uses approximately 33 hours. WM initializes the policy randomly,
    while WM FT fine-tunes the data-collection policy inside the same
    learned world model.}
    \label{fig:sim2sim_data_efficiency}
\end{figure}

%% file: figures/camera_ready/appendix_knee_position.tex
\begin{figure}[htbp]
    \centering
    \includegraphics[width=0.65\linewidth]{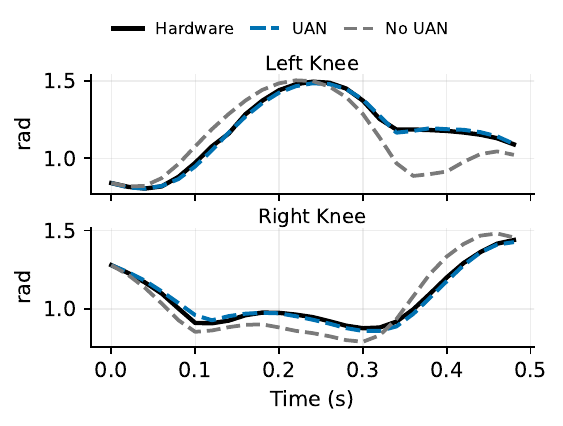}
    \caption{\textbf{Hardware replay knee-position trajectories.} Representative knee-position trajectories on held-out running data compare recorded hardware motion with nominal and UAN-augmented simulator replay of identical recorded commands from the same relabeled initial states.}
    \label{fig:knee_position_replay}
\end{figure}

%% file: corl_2026_template_submission/reference.bib
@misc{schulman2017proximal,
      title={Proximal Policy Optimization Algorithms}, 
      author={John Schulman and Filip Wolski and Prafulla Dhariwal and Alec Radford and Oleg Klimov},
      year={2017},
      eprint={1707.06347},
      archivePrefix={arXiv},
      primaryClass={cs.LG}
}

@misc{NVIDIA_Isaac_Sim,
author = {{NVIDIA}},
license = {Apache-2.0},
title = {{Isaac Sim}},
url = {https://github.com/isaac-sim/IsaacSim},
version = {5.1.0}
}

@misc{shin2023actuatorconstrained,
      title={Actuator-Constrained Reinforcement Learning for High-Speed Quadrupedal Locomotion}, 
      author={Young-Ha Shin and Tae-Gyu Song and Gwanghyeon Ji and Hae-Won Park},
      year={2023},
      eprint={2312.17507},
      archivePrefix={arXiv},
      primaryClass={cs.RO}
}

@misc{tobin2017domain,
      title={Domain Randomization for Transferring Deep Neural Networks from Simulation to the Real World}, 
      author={Josh Tobin and Rachel Fong and Alex Ray and Jonas Schneider and Wojciech Zaremba and Pieter Abbeel},
      year={2017},
      eprint={1703.06907},
      archivePrefix={arXiv},
      primaryClass={cs.RO}
}

@inproceedings{tan2018simtoreal,
      title={Sim-to-Real: Learning Agile Locomotion For Quadruped Robots}, 
      author={Jie Tan and Tingnan Zhang and Erwin Coumans and Atil Iscen and Yunfei Bai and Danijar Hafner and Steven Bohez and Vincent Vanhoucke},
      year={2018},
      booktitle={Robotics: Science and Systems (RSS)}
}

@inproceedings{rudin2022learningwalkminutesusing,
      title={Learning to Walk in Minutes Using Massively Parallel Deep Reinforcement Learning}, 
      author={Nikita Rudin and David Hoeller and Philipp Reist and Marco Hutter},
      year={2021},
      booktitle={Proceedings of the Conference on Robot Learning (CoRL)}
}

@inproceedings{kumar2021rmarapidmotoradaptation,
      title={RMA: Rapid Motor Adaptation for Legged Robots}, 
      author={Ashish Kumar and Zipeng Fu and Deepak Pathak and Jitendra Malik},
      year={2021},
      booktitle={Robotics: Science and Systems (RSS)}
}

@misc{long2024hybridinternalmodellearning,
      title={Hybrid Internal Model: Learning Agile Legged Locomotion with Simulated Robot Response}, 
      author={Junfeng Long and Zirui Wang and Quanyi Li and Jiawei Gao and Liu Cao and Jiangmiao Pang},
      year={2024},
      eprint={2312.11460},
      archivePrefix={arXiv},
      primaryClass={cs.RO},
      url={https://arxiv.org/abs/2312.11460}, 
}

@article{Hwangbo_2019,
   title={Learning agile and dynamic motor skills for legged robots},
   volume={4},
   ISSN={2470-9476},
   url={http://dx.doi.org/10.1126/scirobotics.aau5872},
   DOI={10.1126/scirobotics.aau5872},
   number={26},
   journal={Science Robotics},
   publisher={American Association for the Advancement of Science (AAAS)},
   author={Hwangbo, Jemin and Lee, Joonho and Dosovitskiy, Alexey and Bellicoso, Dario and Tsounis, Vassilios and Koltun, Vladlen and Hutter, Marco},
   year={2019},
   month=jan }

@INPROCEEDINGS{SchwendemanHumanoidSysID,
  author={Schwendeman, Laura and SaLoutos, Andrew and Stanger-Jones, Elijah and Kim, Sangbae},
  booktitle={2023 IEEE-RAS 22nd International Conference on Humanoid Robots (Humanoids)}, 
  title={Improving Domain Transfer of Robot Dynamics Models with Geometric System Identification and Learned Friction Compensation}, 
  year={2023},
  volume={},
  number={},
  pages={1-8},
  doi={10.1109/Humanoids57100.2023.10375233}}

@article{Lee_2020,
   title={Learning quadrupedal locomotion over challenging terrain},
   volume={5},
   ISSN={2470-9476},
   url={http://dx.doi.org/10.1126/scirobotics.abc5986},
   DOI={10.1126/scirobotics.abc5986},
   number={47},
   journal={Science Robotics},
   publisher={American Association for the Advancement of Science (AAAS)},
   author={Lee, Joonho and Hwangbo, Jemin and Wellhausen, Lorenz and Koltun, Vladlen and Hutter, Marco},
   year={2020},
   month=oct }

@article{AtkesonEstimation,
author = {Christopher G. Atkeson and Chae H. An and John M. Hollerbach},
title ={Estimation of Inertial Parameters of Manipulator Loads and Links},
journal = {The International Journal of Robotics Research},
volume = {5},
number = {3},
pages = {101-119},
year = {1986},
doi = {10.1177/027836498600500306},
URL = {        https://doi.org/10.1177/027836498600500306 },
eprint = {        https://doi.org/10.1177/027836498600500306 }
}

@ARTICLE{LeeGeometric,
  author={Lee, Taeyoon and Wensing, Patrick M. and Park, Frank C.},
  journal={IEEE Transactions on Robotics}, 
  title={Geometric Robot Dynamic Identification: A Convex Programming Approach}, 
  year={2020},
  volume={36},
  number={2},
  pages={348-365},
  doi={10.1109/TRO.2019.2926491}}

@article{WensingGeometricObservability,
author = {Patrick M. Wensing and Günter Niemeyer and Jean-Jacques E. Slotine},
title ={A geometric characterization of observability in inertial parameter identification},
journal = {The International Journal of Robotics Research},
volume = {43},
number = {14},
pages = {2274-2302},
year = {2024},
doi = {10.1177/02783649241258215},
URL = {        https://doi.org/10.1177/02783649241258215 },
eprint = {        https://doi.org/10.1177/02783649241258215 }
}

@INPROCEEDINGS{mujoco,
  author={Todorov, Emanuel and Erez, Tom and Tassa, Yuval},
  booktitle={2012 IEEE/RSJ International Conference on Intelligent Robots and Systems}, 
  title={MuJoCo: A physics engine for model-based control}, 
  year={2012},
  volume={},
  number={},
  pages={5026-5033},
  doi={10.1109/IROS.2012.6386109}}

@inproceedings{gruenstein2021residual,
  title={Residual model learning for microrobot control},
  author={Gruenstein, Joshua and Chen, Tao and Doshi, Neel and Agrawal, Pulkit},
  booktitle={2021 IEEE International Conference on Robotics and Automation (ICRA)},
  pages={7219--7226},
  year={2021},
  organization={IEEE}
}

@misc{margolis2022rapidlocomotionreinforcementlearning,
      title={Rapid Locomotion via Reinforcement Learning}, 
      author={Gabriel B Margolis and Ge Yang and Kartik Paigwar and Tao Chen and Pulkit Agrawal},
      year={2022},
      eprint={2205.02824},
      archivePrefix={arXiv},
      primaryClass={cs.RO},
      url={https://arxiv.org/abs/2205.02824}, 
}

@inproceedings{sontakke2023residual,
  title={Residual physics learning and system identification for sim-to-real transfer of policies on buoyancy assisted legged robots},
  author={Sontakke, Nitish and Chae, Hosik and Lee, Sangjoon and Huang, Tianle and Hong, Dennis W and Hal, Sehoon},
  booktitle={2023 IEEE/RSJ International Conference on Intelligent Robots and Systems (IROS)},
  pages={392--399},
  year={2023},
  organization={IEEE}
}

@article{zeng2020tossingbot,
  title={Tossingbot: Learning to throw arbitrary objects with residual physics},
  author={Zeng, Andy and Song, Shuran and Lee, Johnny and Rodriguez, Alberto and Funkhouser, Thomas},
  journal={IEEE Transactions on Robotics},
  volume={36},
  number={4},
  pages={1307--1319},
  year={2020},
  publisher={IEEE}
}

@inproceedings{nahrendra2023dreamwaq,
  title={DreamWaQ: Learning robust quadrupedal locomotion with implicit terrain imagination via deep reinforcement learning},
  author={Nahrendra, I Made Aswin and Yu, Byeongho and Myung, Hyun},
  booktitle={2023 IEEE International Conference on Robotics and Automation (ICRA)},
  pages={5078--5084},
  year={2023},
  organization={IEEE}
}

@article{ji2022concurrent,
  title={Concurrent training of a control policy and a state estimator for dynamic and robust legged locomotion},
  author={Ji, Gwanghyeon and Mun, Juhyeok and Kim, Hyeongjun and Hwangbo, Jemin},
  journal={IEEE Robotics and Automation Letters},
  volume={7},
  number={2},
  pages={4630--4637},
  year={2022},
  publisher={IEEE}
}

@inproceedings{khosla1985parameter,
  title={Parameter identification of robot dynamics},
  author={Khosla, Pradeep K and Kanade, Takeo},
  booktitle={1985 24th IEEE conference on decision and control},
  pages={1754--1760},
  year={1985},
  organization={IEEE}
}

@inproceedings{sobanbabu2025sampling,
  title={Sampling-based System Identification with Active Exploration for Legged Sim2Real Learning},
  author={Sobanbabu, Nikhil and He, Guanqi and He, Tairan and Yang, Yuxiang and Shi, Guanya},
  booktitle={9th Annual Conference on Robot Learning},
  year={2025}
}

@article{bjelonic2025towards,
  title         = {Towards Bridging the Gap: Systematic Sim-to-Real Transfer for Diverse Legged Robots},
  author        = {Bjelonic, Filip and Tischhauser, Fabian and Hutter, Marco},
  journal       = {arXiv preprint arXiv:2509.06342},
  year          = {2025},
  eprint        = {2509.06342},
  archivePrefix = {arXiv},
  primaryClass  = {cs.RO},
}

@article{le2021differentiable,
  title={Differentiable simulation for physical system identification},
  author={Le Lidec, Quentin and Kalevatykh, Igor and Laptev, Ivan and Schmid, Cordelia and Carpentier, Justin},
  journal={IEEE Robotics and Automation Letters},
  volume={6},
  number={2},
  pages={3413--3420},
  year={2021},
  publisher={IEEE}
}

@article{liao2025beyondmimic,
  title={Beyondmimic: From motion tracking to versatile humanoid control via guided diffusion},
  author={Liao, Qiayuan and Truong, Takara E and Huang, Xiaoyu and Gao, Yuman and Tevet, Guy and Sreenath, Koushil and Liu, C Karen},
  journal={arXiv preprint arXiv:2508.08241},
  year={2025}
}

@article{li2025robotic,
  title={Robotic world model: A neural network simulator for robust policy optimization in robotics},
  author={Li, Chenhao and Krause, Andreas and Hutter, Marco},
  journal={arXiv preprint arXiv:2501.10100},
  year={2025}
}

@article{li2025uncertainty,
  title={Uncertainty-Aware Robotic World Model Makes Offline Model-Based Reinforcement Learning Work on Real Robots},
  author={Li, Chenhao and Krause, Andreas and Hutter, Marco},
  journal={arXiv preprint arXiv:2504.16680},
  year={2025}
}

@inproceedings{he2025asap,
  title={ASAP: Aligning Simulation and Real-World Physics for Learning Agile Humanoid Whole-Body Skills},
  author={He, Tairan and Gao, Jiawei and Xiao, Wenli and Zhang, Yuanhang and Wang, Zi and Wang, Jiashun and Luo, Zhengyi and He, Guanqi and Sobanbab, Nikhil and Pan, Chaoyi and others},
  booktitle={Robotics: Science and Systems (RSS)},
  year={2025}
}

@inproceedings{
fey2025bridging,
title={Bridging the Sim-to-Real Gap for Athletic Loco-Manipulation},
author={Nolan Fey and Gabriel B. Margolis and Martin Peticco and Pulkit Agrawal},
booktitle={Robotics: Science and Systems (RSS)},
year={2025},
url={https://openreview.net/forum?id=0HwPNCqQft}
}

@inproceedings{margolis2023walk,
  title={Walk these ways: Tuning robot control for generalization with multiplicity of behavior},
  author={Margolis, Gabriel B and Agrawal, Pulkit},
  booktitle={Conference on Robot Learning},
  pages={22--31},
  year={2023},
  organization={PMLR}
}

@inproceedings{portela2024learning,
  title={Learning force control for legged manipulation},
  author={Portela, Tifanny and Margolis, Gabriel B and Ji, Yandong and Agrawal, Pulkit},
  booktitle={2024 IEEE International Conference on Robotics and Automation (ICRA)},
  pages={15366--15372},
  year={2024},
  organization={IEEE}
}

@article{sleiman2026zest,
  title={ZEST: Zero-shot Embodied Skill Transfer for Athletic Robot Control},
  author={Sleiman, Jean Pierre and Li, He and Adu-Bredu, Alphonsus and Deits, Robin and Kumar, Arun and Bergamin, Kevin and Bhardwaj, Mohak and Biddlestone, Scott and Burger, Nicola and Estrada, Matthew A and others},
  journal={arXiv preprint arXiv:2602.00401},
  year={2026}
}

@article{margolis2025softmimic,
  title={Softmimic: Learning compliant whole-body control from examples},
  author={Margolis, Gabriel B and Wang, Michelle and Fey, Nolan and Agrawal, Pulkit},
  journal={arXiv preprint arXiv:2510.17792},
  year={2025}
}

@article{hou2026world,
  title={World Model for Robot Learning: A Comprehensive Survey},
  author={Hou, Bohan and Li, Gen and Jia, Jindou and An, Tuo and Guo, Xinying and Leng, Sicong and Geng, Haoran and Ze, Yanjie and Harada, Tatsuya and Torr, Philip and others},
  journal={arXiv preprint arXiv:2605.00080},
  year={2026}
}

@article{cha2025sim,
  title={Sim-to-real of humanoid locomotion policies via joint torque space perturbation injection},
  author={Cha, Junhyeok Rui and Cha, Woohyun and Shin, Jaeyong and Kim, Donghyeon and Park, Jaeheung},
  journal={arXiv preprint arXiv:2504.06585},
  year={2025}
}

@inproceedings{hansen2025hierarchical,
  title={Hierarchical world models as visual whole-body humanoid controllers},
  author={Hansen, Nick and SV, Jyothir and Sobal, Vlad and LeCun, Yann and Wang, Xiaolong and Su, Hao},
  booktitle={International Conference on Learning Representations},
  volume={2025},
  pages={62175--62195},
  year={2025}
}

@inproceedings{wu2023daydreamer,
  title={Daydreamer: World models for physical robot learning},
  author={Wu, Philipp and Escontrela, Alejandro and Hafner, Danijar and Abbeel, Pieter and Goldberg, Ken},
  booktitle={Conference on robot learning},
  pages={2226--2240},
  year={2023},
  organization={PMLR}
}

@inproceedings{xie2021dynamics,
  title={Dynamics randomization revisited: A case study for quadrupedal locomotion},
  author={Xie, Zhaoming and Da, Xingye and Van de Panne, Michiel and Babich, Buck and Garg, Animesh},
  booktitle={2021 IEEE International Conference on Robotics and Automation (ICRA)},
  pages={4955--4961},
  year={2021},
  organization={IEEE}
}

@article{muratore2022robot,
  title={Robot learning from randomized simulations: A review},
  author={Muratore, Fabio and Ramos, Fabio and Turk, Greg and Yu, Wenhao and Gienger, Michael and Peters, Jan},
  journal={Frontiers in Robotics and AI},
  volume={9},
  pages={799893},
  year={2022},
  publisher={Frontiers Media SA}
}

@inproceedings{campanaro2024learning,
  title={Learning and deploying robust locomotion policies with minimal dynamics randomization},
  author={Campanaro, Luigi and Gangapurwala, Siddhant and Merkt, Wolfgang and Havoutis, Ioannis},
  booktitle={6th Annual Learning for Dynamics \& Control Conference},
  pages={578--590},
  year={2024},
  organization={PMLR}
}

@inproceedings{liu2025locoformer,
  title={LocoFormer: Generalist Locomotion via Long-context Adaptation},
  author={Liu, Min and Pathak, Deepak and Agarwal, Ananye},
  booktitle={Conference on Robot Learning},
  pages={532--546},
  year={2025},
  organization={PMLR}
}

@inproceedings{jiang2021simgan,
  title={Simgan: Hybrid simulator identification for domain adaptation via adversarial reinforcement learning},
  author={Jiang, Yifeng and Zhang, Tingnan and Ho, Daniel and Bai, Yunfei and Liu, C Karen and Levine, Sergey and Tan, Jie},
  booktitle={2021 IEEE International Conference on Robotics and Automation (ICRA)},
  pages={2884--2890},
  year={2021},
  organization={IEEE}
}

@inproceedings{golemo2018sim,
  title={Sim-to-real transfer with neural-augmented robot simulation},
  author={Golemo, Florian and Taiga, Adrien Ali and Courville, Aaron and Oudeyer, Pierre-Yves},
  booktitle={Conference on Robot Learning},
  pages={817--828},
  year={2018},
  organization={PMLR}
}

@article{yu2025neural,
  title={Neural Fidelity Calibration for Informative Sim-to-Real Adaptation},
  author={Yu, Youwei and Liu, Lantao},
  journal={arXiv preprint arXiv:2504.08604},
  year={2025}
}

@article{gao2024sim,
  title={Sim-to-real of soft robots with learned residual physics},
  author={Gao, Junpeng and Michelis, Mike Y and Spielberg, Andrew and Katzschmann, Robert K},
  journal={IEEE Robotics and Automation Letters},
  volume={9},
  number={10},
  pages={8523--8530},
  year={2024},
  publisher={IEEE}
}

@article{sun2026mosaic,
  title={Mosaic: Bridging the sim-to-real gap in generalist humanoid motion tracking and teleoperation with rapid residual adaptation},
  author={Sun, Zhenguo and Huang, Bo-Sheng and Peng, Yibo and Li, Xukun and Ma, Jingyu and Sun, Yu and Li, Zhe and Jiang, Haojun and Gao, Biao and Bing, Zhenshan and others},
  journal={arXiv preprint arXiv:2602.08594},
  year={2026}
}

@article{sutton1991dyna,
  title={Dyna, an integrated architecture for learning, planning, and reacting},
  author={Sutton, Richard S},
  journal={ACM Sigart Bulletin},
  volume={2},
  number={4},
  pages={160--163},
  year={1991},
  publisher={ACM New York, NY, USA}
}

@article{zakka2026mjlab,
  title={mjlab: A Lightweight Framework for GPU-Accelerated Robot Learning},
  author={Zakka, Kevin and Liao, Qiayuan and Yi, Brent and Lay, Louis Le and Sreenath, Koushil and Abbeel, Pieter},
  journal={arXiv preprint arXiv:2601.22074},
  year={2026}
}

@article{muller2026olaf,
  title={Olaf: Bringing an animated character to life in the physical world},
  author={M{\"u}ller, David and Knoop, Espen and Mylonopoulos, Dario and Serifi, Agon and Hopkins, Michael A and Grandia, Ruben and B{\"a}cher, Moritz},
  journal={IEEE Robotics and Automation Letters},
  year={2026},
  publisher={IEEE}
}

@misc{miller2025highperformancereinforcementlearningspot,
      title={High-Performance Reinforcement Learning on Spot: Optimizing Simulation Parameters with Distributional Measures}, 
      author={AJ Miller and Fangzhou Yu and Michael Brauckmann and Farbod Farshidian},
      year={2025},
      eprint={2504.17857},
      archivePrefix={arXiv},
      primaryClass={cs.LG},
      url={https://arxiv.org/abs/2504.17857}, 
}

@misc{cmumocap,
  title={{CMU} Graphics Lab Motion Capture Database},
  author={{Carnegie Mellon University}},
  howpublished={\url{http://mocap.cs.cmu.edu}},
}

@article{schwarke2025rslrl,
  title={RSL-RL: A Learning Library for Robotics Research},
  author={Schwarke, Clemens and Mittal, Mayank and Rudin, Nikita and Hoeller, David and Hutter, Marco},
  journal={arXiv preprint arXiv:2509.10771},
  year={2025}
}

@misc{wu2026perceptivehumanoidparkourchaining,
      title={Perceptive Humanoid Parkour: Chaining Dynamic Human Skills via Motion Matching}, 
      author={Zhen Wu and Xiaoyu Huang and Lujie Yang and Yuanhang Zhang and Xi Chen and Pieter Abbeel and Rocky Duan and Angjoo Kanazawa and Carmelo Sferrazza and Guanya Shi and C. Karen Liu},
      year={2026},
      eprint={2602.15827},
      archivePrefix={arXiv},
      primaryClass={cs.RO},
      url={https://arxiv.org/abs/2602.15827}, 
}

@misc{margolis2023learningphysicalpropertiesactive,
      title={Learning to See Physical Properties with Active Sensing Motor Policies}, 
      author={Gabriel B. Margolis and Xiang Fu and Yandong Ji and Pulkit Agrawal},
      year={2023},
      eprint={2311.01405},
      archivePrefix={arXiv},
      primaryClass={cs.RO},
      url={https://arxiv.org/abs/2311.01405}, 
}
